\documentclass{article}

\usepackage[main, final]{neurips_2026}
\usepackage[utf8]{inputenc}
\usepackage[T1]{fontenc}
\usepackage{hyperref}
\usepackage{url}
\usepackage{booktabs}
\usepackage{amsfonts}
\usepackage{amsmath}
\usepackage{amssymb}
\usepackage{nicefrac}
\usepackage{microtype}
\usepackage{xcolor}
\usepackage{graphicx}
\usepackage{algorithm}
\usepackage{subcaption}
\usepackage{multirow}

\usepackage{tabularx}
\usepackage{makecell}
\usepackage{threeparttable}
\usepackage{colortbl}
\usepackage{lipsum}
\usepackage{placeins}
\usepackage{wrapfig}
\usepackage{algpseudocode}

\title{Learning-to-Memorize: Dynamic Context Management for Long-Horizon Autoregressive Video Generation}

\author{Haowei Zhu\textsuperscript{\rm 1,2}, Qijie Wang\textsuperscript{\rm 1}, Jia Li\textsuperscript{\rm 2}, Xing Wang\textsuperscript{\rm 2}, Tianyu Zhao\textsuperscript{\rm 2}, Jiexi Wang\textsuperscript{\rm 2}, \\\textbf{Xurui Peng\textsuperscript{\rm 2}, Xinglong Wu\textsuperscript{\rm 2}, Junhai Yong\textsuperscript{\rm 1}, Bin Wang\textsuperscript{\rm 1}\thanks{Corresponding author: {wangbins@tsinghua.edu.cn}}}\\
\textsuperscript{\rm 1} Tsinghua University
\textsuperscript{\rm 2} ByteDance Seed}

\begin{document}

\maketitle

\begin{abstract}

Autoregressive video generation has achieved impressive results on short clips, yet generating long videos remains challenging due to error accumulation over extended horizons, where each predicted frame depends on potentially imperfect previous frames. Existing approaches typically rely on static context management strategies, such as sliding-window caches that discard early frames or fixed sink frames that permanently preserve initial content. These designs either lose critical long-range context or anchor generation to outdated information, which can lead to degraded fidelity, motion stagnation, and reduced diversity.
In this paper, we propose \textbf{Learning to Memorize (L2M)}, a learned context memory management framework that dynamically retains, evicts, and evolves historical frames according to predicted importance. The framework consists of an \emph{importance prediction router} that estimates the relevance of each historical frame, \emph{degradation-aware training} that improves robustness to noisy or corrupted context, and a \emph{dynamic memory initialization and evolution} mechanism that updates long-term memory using exponential moving average (EMA) importance scores. This allows high-quality frames to replace stale anchors and enables memory to adapt naturally to scene evolution. Trained with a teacher-student distillation objective and sparsity regularization, L2M makes efficient use of a fixed KV-cache budget while preserving the most informative contextual information.
Extensive experiments demonstrate that L2M improves long-horizon video generation quality and temporal consistency over existing baselines, providing a learned strategy for context memory management in long autoregressive video generation.

\end{abstract}

\section{Introduction}
\label{sec:intro}
Recent advances in video diffusion models~\cite{yang2024cogvideox, wan2025, HaCohen2024LTXVideo, ho2022video, blattmann2023align, polyak2024movie} have demonstrated remarkable capabilities in synthesizing short video clips (e.g., 5--10 seconds) with both high visual fidelity and coherent motion dynamics.
Building upon this progress, emerging interactive applications such as world models~\cite{rre, yan} and real-time content creation~\cite{shin2025motionstream} increasingly demand \emph{long-horizon} autoregressive video generation, where frames are streamed sequentially over extended durations (e.g., 1--2 minutes).

Self Forcing~\cite{huang2025self} and its variants~\cite{yang2025longlive, liu2025rolling, cui2025self} have become standard paradigms in autoregressive video generation, leveraging causal attention masks and Key-Value (KV) caches from previously generated frames.
However, this autoregressive formulation is inherently susceptible to \emph{error accumulation over long horizons}~\cite{yin2025slow, huang2025self, wang2025error}, depends on previously generated frames that may contain errors, leading to progressive fidelity degradation, color drift, and texture blurring.

To mitigate error accumulation, recent methods have explored two principal directions.
The first direction introduces history corruption during training~\cite{chen2024diffusion, song2025history}, adding noise to previous frames to improve robustness.
While effective to some degree, a discrepancy remains between artificially injected noise and the actual distribution of generated errors, leaving models vulnerable to long-horizon drift.
The second direction, inspired by the attention sink mechanism from large language models (LLMs)~\cite{xiao2023efficient}, retains a small set of initial tokens as permanent anchors to stabilize the attention distribution~\cite{yesiltepe2025infinity}.
Deep Forcing~\cite{yi2025deep} extends this idea with deep sinks and participative compression, while LongLive~\cite{yang2025longlive} and Rolling Forcing~\cite{liu2025rolling} incorporate sink-aware training.
PackForcing~\cite{mao2026packforcing} introduces a three-partition KV cache with token compression for bounded memory.

Despite these advances, we identify a fundamental limitation shared by existing methods: their context management strategies are \textbf{static} and \textbf{rule-based}. Specifically, (1) FIFO policies discard frames solely based on recency, without considering their semantic importance; (2) fixed sink strategies permanently retain initial frames, regardless of whether they contain useful information, such as during transitions or occlusions; and (3) in long-horizon dynamic scenes, early sink frames often become outdated as subsequent content diverges significantly, reducing their relevance as anchors.
Recent methods \cite{yi2025deep, mao2026packforcing, zhang2026frame} that select tokens based on similarity are also suboptimal, as they tend to prioritize recently generated frames, which are naturally more similar to each other, rather than identifying truly informative historical content.

In this paper, we propose \textbf{Learning-to-Memorize (L2M)}, a learned context management framework that dynamically determines which historical frames should be retained, evicted, or promoted during long-horizon autoregressive video generation. Our key insight is that context management should be content-adaptive rather than governed by fixed rules: the model learns to predict which historical frames are most valuable for future generation and manages the limited KV-cache budget accordingly.

The framework consists of three main components. First, an \emph{importance prediction router} evaluates the relevance of each historical frame relative to the current frame through cross-frame feature interactions, predicting an importance score of each context frames. Second, a \emph{degradation-aware training} scheme augments historical context with simulated degradations such as noise, blur, or occlusion, teaching the router to assign lower importance to unreliable frames and improving inference-time robustness. Third, a \emph{dynamic memory evolving} mechanism maintains exponential moving average (EMA) importance scores and promotes high-quality newly generated frames into the long-term memory sink set while replacing stale anchors during sampling, allowing the memory to evolve naturally with the video content.
The entire framework is trained with a teacher-student distillation objective, where a teacher operating on full context provides supervision for a student operating on router-pruned context, combined with sparsity regularization to encourage efficient cache utilization.
Our contributions are summarized as follows:
\begin{itemize}
    \item We identify a fundamental limitation of static context management in autoregressive video generation and propose a \emph{learned} context management framework, \textbf{L2M}, which dynamically retains, evicts, and updates historical frames based on predicted importance.
    
    \item We introduce a lightweight importance prediction router together with degradation-aware training via context perturbation and a dynamic memory mechanism that initializes and continuously evolves long-term context. These components jointly enable content-adaptive memory management in long video generation.
    
    \item Extensive experiments demonstrate that L2M consistently outperforms existing baselines on long-horizon video generation, improving temporal consistency and visual fidelity, and establishing a new learned context management method in autoregressive video generation.
\end{itemize}

\section{Related Work}
\label{sec:related}

\noindent\textbf{Video Diffusion and Long Video Generation.}  
Early video generation extended 2D image diffusion models with temporal modules~\cite{ho2022video, blattmann2023align, singer2022make}, while recent Diffusion Transformers (DiTs)~\cite{peebles2023scalable, brooks2024video} treat videos as spatiotemporal patches for scalable 3D attention. State-of-the-art models such as CogVideoX~\cite{yang2024cogvideox}, MovieGen~\cite{polyak2024movie}, Wan~\cite{wan2025}, and LTX-Video~\cite{HaCohen2024LTXVideo} produce high-fidelity short clips, and flow-matching techniques~\cite{lipman2023flow, liu2022flow} accelerate convergence. However, joint spatiotemporal modeling for minute-level videos remains computationally intractable, motivating memory-efficient autoregressive strategies~\cite{qiu2024freenoisetuning, ge2023preserve, hong2023large, henschel2024streamingt2v} that either rely on hierarchical planning or multi-stage pipelines, often at the cost of complexity.

\noindent\textbf{Autoregressive Video Generation and Context Management.}  
Autoregressive approaches synthesize frames sequentially using KV caches, with methods like CausVid~\cite{yin2025slow}, Diffusion Forcing~\cite{chen2024diffusion}, and Self Forcing~\cite{huang2025self} bridging the train-inference gap or improving robustness~\cite{song2025history}. Extensions such as Rolling Forcing~\cite{liu2025rolling}, LongLive~\cite{yang2025longlive}, and Self Forcing++~\cite{cui2025self} enable minute-scale generation while maintaining visual continuity. 
Despite these advances, existing methods largely rely on hand-crafted cache-management rules, including FIFO eviction, fixed sink retention, and attention-based dynamic selection. L2M learns a frame-importance predictor and additionally updates long-term memory.

\noindent\textbf{Attention Sinks and KV Cache Optimization.}  
The attention sink phenomenon, first observed in large language models~\cite{xiao2023efficient}, concentrates attention on initial tokens, and retaining these sinks in a sliding window stabilizes long-context generation. Various KV cache compression methods have been proposed for efficiency, including H2O~\cite{zhang2023h2o}, SnapKV~\cite{li2024snapkv}, LaCache~\cite{shilacache}, and D2O~\cite{Ghadia2025DialogueWL}. Video generation methods~\cite{yi2025deep, mao2026packforcing, li2026train, fu2026towards} extend sinks with deep tokens or spatiotemporal compression. All these approaches rely on static or heuristic criteria, whereas L2M learns frame importance via a trainable router and dynamically evolves long-term memory, achieving content-adaptive context management within a single-stage causal framework while maintaining visual fidelity and natural content evolution.

\section{Method}
\label{sec:method}

\subsection{Overview}
\label{sec:overview}

We propose a learned context memory management framework for long-horizon autoregressive video generation.
Given an autoregressive video generation model that generates the current latent chunk $z_b$ conditioned on historical context $\mathcal{C}_{b}$, our goal is to learn a context selection policy
\begin{equation}
\widetilde{\mathcal{C}}_{b}
=
\pi(z_b,\mathcal{C}_{b}),
\quad
|\widetilde{\mathcal{C}}_{b}|\leq K,
\label{eq:policy}
\end{equation}
where $b$ indexes the current generation chunk and $K$ denotes the cache budget measured in latent frames.
Unlike prior approaches that define $\pi$ through static rules (e.g., FIFO eviction or fixed sinks), we \emph{learn} $\pi$ from data such that it can dynamically decide which historical frames to retain, evict, and promote based on their content and predicted utility for future generation.

\begin{figure*}[t]
  \centering
  \includegraphics[width=0.95\textwidth]{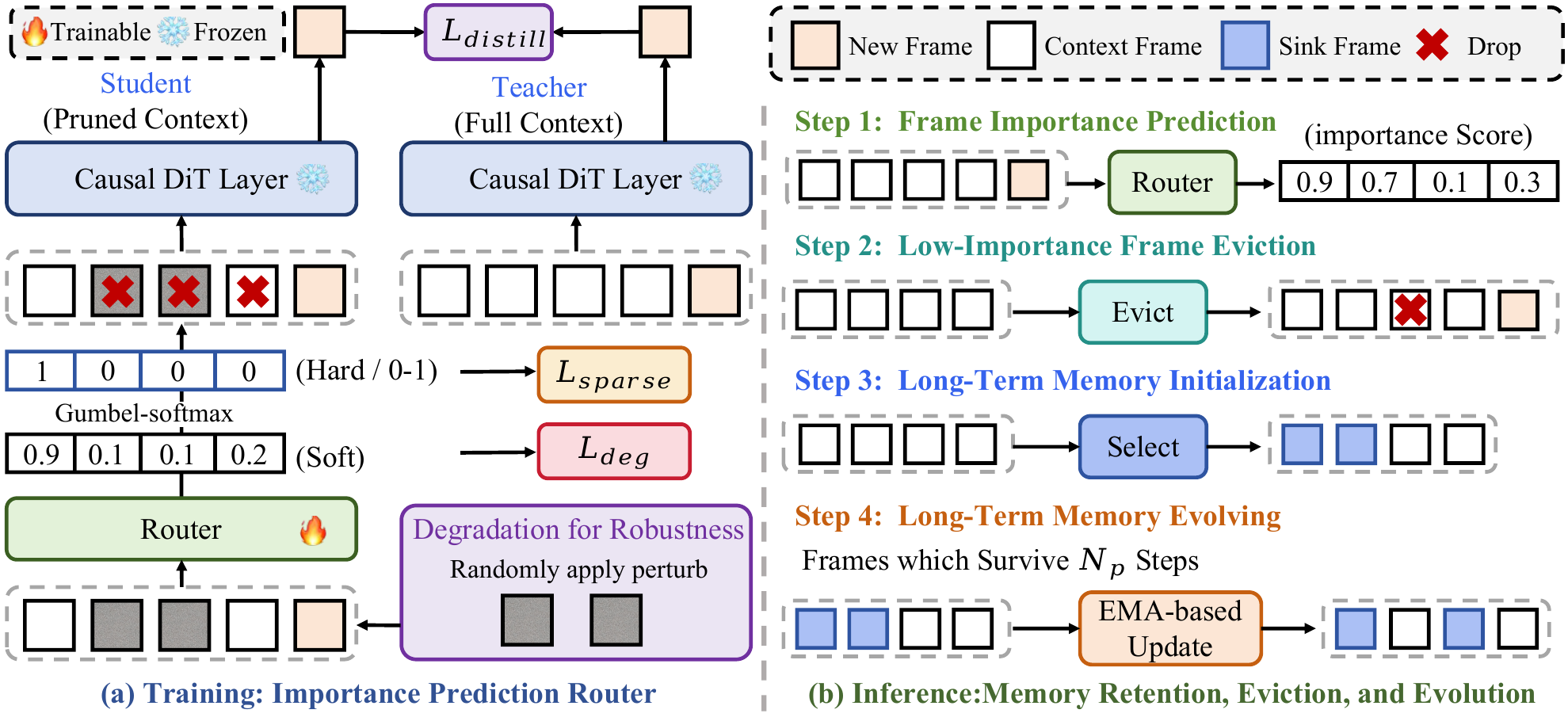}
\caption{
Overview of L2M. \textbf{(a) Training phase}: Given the current chunk feature $\bar{f}_b$ and historical latent frame features $\{\bar{f}_i\}_{i<(b-1)B_f+1}$, the importance-aware frame router predicts importance scores for each historical latent frame. Binary frame selection masks $m_i\in\{0,1\}$ are obtained via Gumbel-Softmax discretization. The pruned context is fed to the student network, which is supervised by the teacher operating on the full context. Degradation-aware training augments a subset of historical frames with degradations, encouraging the router to assign them lower retention probabilities. 
\textbf{(b) Inference phase}: The router predicts frame importance scores, and low-scoring latent frames are evicted to meet the cache budget. High-scoring frames are dynamically selected as long-term memory and periodically updated using maintained EMA scores, continuously promoting informative historical frames.
}
  \label{fig:framework}
  \vspace{-5pt}
\end{figure*}

Our framework consists of three core components: (1)~an \emph{importance prediction router} that predicts the importance of each historical frame (Section~\ref{sec:router}); (2)~\emph{degradation-aware training} that simulates context degradation to improve router robustness (Section~\ref{sec:robust}); and (3)~\emph{dynamic memory management mechanism} that maintains an evolving long-term memory set at inference time (Section~\ref{sec:dynamic_memory}).
The overall framework is illustrated in Figure~\ref{fig:framework}.

\subsection{Preliminaries: Autoregressive Video Generation}
\label{sec:prelim}
We build upon the autoregressive video diffusion framework~\cite{huang2025self, yin2025slow}. A video consisting of $T$ latent frames is divided into $N_c=\lceil T/B_f\rceil$ non-overlapping chunks, each containing $B_f$ latent frames (typically $B_f=3$). We use $b\in\{1,\ldots,N_c\}$ to index chunks and $i\in\{1,\ldots,T\}$ to index individual latent frames. Specifically, the $b$-th chunk is denoted by $z_b=[x_{(b-1)B_f+1},\ldots,x_{(bB_f,T)}]\in\mathbb{R}^{B_f\times C\times H\times W}$, where $x_i\in\mathbb{R}^{C\times H\times W}$ denotes the $i$-th latent frame.
During the generation of chunk $b$, each transformer layer $l$ attends to the KV pairs cached from all previously generated latent frames:

\begin{equation}
\mathcal{C}_b^l =
\left\{
(K_i^l,V_i^l)
\right\}_{i=1}^{(b-1)B_f},
\end{equation}

For each latent frame $i$, the cached KV tensors at transformer layer $l$ are denoted by $K_i^l, V_i^l \in \mathbb{R}^{hw \times N_h \times d_h},$ where $h$ and $w$ denote the spatial token dimensions of each latent frame, $N_h$ is the number of attention heads, and $d_h$ is the per-head dimension. We omit the batch dimension for simplicity. The KV cache of chunk $b$ is obtained by concatenating the KV tensors of its constituent latent frames along the token dimension.

As generation proceeds, the KV cache grows linearly.
For minute-long videos, this quickly exceeds available GPU memory, necessitating a context management strategy to bound the cache size.
Prior approaches employ either FIFO eviction~\cite{huang2025self, yin2025slow}, where the oldest chunk is discarded when the cache is full, or attention sinks~\cite{yi2025deep, yang2025longlive}, where initial chunks are permanently retained.
Our method instead \emph{learns} which frames to retain.

\subsection{Importance-aware Frame Router}
\label{sec:router}

The core component of our framework is a lightweight frame router that evaluates the importance of each historical frame for generating the current chunk. 
Let $\bar{f}_b$ denote the average-pooled feature of the current chunk $z_b$ along the temporal dimension. For each historical latent frame $x_i$, its corresponding feature $f_i$ is averaged along the spatial token dimension to obtain $\bar{f}_i$.
To capture content-dependent relevance, we first concatenate $\bar{f}_t$ and $\bar{f}_i$ along the channel dimension and pass the result through a lightweight multi-layer perceptron (MLP). The router then produces the importance scores:
\begin{equation}
s_i = \operatorname{Router}(\bar{f}_b,\bar{f}_i)
= \operatorname{Softmax}
\left(
\operatorname{MLP}
([\bar{f}_b \oplus \bar{f}_i])
\right)
\in \mathbb{R}^2.
\label{eq:router}
\end{equation}
We use the predicted probability of the first category as the importance score of each historical frame, i.e., $r_i = [s_i]_0$, where a higher score indicates a higher likelihood of retaining the frame.
During training, the Softmax output is combined with the Gumbel-Softmax trick~\citep{jang2017categorical} to produce a differentiable approximation of discrete selection, while at inference time a standard Softmax is applied. After normalization, $m_{i,0}=1$ indicates that the frame is retained, while $m_{i,1}=1$ indicates it is pruned. The predicted variables $m_i$ are then applied to the historical context to produce the pruned context for the student network during training or for memory-efficient inference.

However, directly discarding or zeroing out frames with $m_i=0$ would interfere with batch parallelism and is non-differentiable. To address this, we employ a differentiable attention masking strategy \cite{rao2021dynamicvit} that completely prevents pruned frames from affecting retained tokens while maintaining a fixed token count. 
Detailed description of this masking strategy is provided in Appendix \ref{sec:supp_attn_with_mask}.

\paragraph{Context Pruning}
The frame-level selection masks $m_i$ are applied to the corresponding historical KV pairs to construct the pruned context $\widetilde{\mathcal{C}}_b^l$. Specifically, all spatial tokens belonging to a latent frame are jointly retained or removed, ensuring that frame-level temporal positions remain consistent.
\begin{equation}
\widetilde{\mathcal{C}}_b^l =
\left\{
(K_i^l,V_i^l)
\mid
i < (b-1)B_f+1,\; m_i=1
\right\}.
\label{eq:pruning}
\end{equation}
During training, the pruned context is fed to the student network, while the full context is used by the teacher network for supervision. At inference time, frames with $m_i=0$ are evicted to satisfy the fixed cache budget.

\paragraph{Learning Objective}
We train the router under a teacher–student distillation framework, with the router being the sole trainable parameter of our approach.
The teacher network, operating on the full context $\mathcal{C}_{t-1}$, produces target features $y^*$, while the student network, operating on the pruned context $\tilde{\mathcal{C}}_{t-1}$, produces output features $y$. We apply both distillation and sparsity losses to each transformer layer, and an additional distillation loss is applied to the model's noise predictions $\hat{\epsilon}$.
The distillation loss encourages the student to match the teacher's output:
\begin{equation}
\mathcal{L}_{\mathrm{distill}} = \| y - y^* \|_2^2.
\label{eq:distill_loss}
\end{equation}

A sparsity regularization term is added to encourage the router to retain as few frames as possible while maintaining generation quality:
\begin{equation}
    \mathcal{L}_{\mathrm{sparse}}
    =
    \frac{1}{N_b}
    \sum_{i=1}^{N_b}m_{i,\mathrm{retain}}.
\label{eq:sparse_loss}
\end{equation}
where $N_b$ is the total number of historical frames. These two objectives together encourage the router to prune as many frames as possible while minimizing information loss, ensuring that the student network retains sufficient context for high-quality generation.
To save memory and enable efficient training, gradients are computed only for a single sampled denoising step~\cite{huang2025self} and the last frame chunk.

\paragraph{Temporal RoPE Adjustment} 
The key cache for each frame encodes temporal position information via the use of 3D Rotary Position Embeddings (RoPE) ~\cite{su2024roformer}. When context frames are evicted, gaps in the temporal positions of the remaining keys are introduced, which can disrupt the positional continuity that the Transformer attention mechanism relies on. To address this, we apply a \emph{continuous temporal RoPE adjustment} to the re-encoded frames. Specifically, for each frame, we compute the difference $\Delta_i$ between its original temporal position in the sequence and its new expected position after eviction. Using the RoPE temporal frequency vector $\boldsymbol{\omega}_t$, the temporal component of the key is adjusted as:
\begin{equation}
K_i^{l,\mathrm{(time)}} \leftarrow
K_i^{l,\mathrm{(time)}}\odot
\exp\left(\mathrm{i}\omega_t\Delta_i\right).
\label{eq:time_adjust}
\end{equation}
where $\mathrm{i}$ is the imaginary unit and $\odot$ denotes element-wise multiplication. This operation rotates $K_i$ in the complex plane to align the relative temporal positions of sink and tail tokens, preserving the positional continuity required for accurate attention computation. This multiplicative adjustment is computationally efficient, has been widely adopted in prior works \cite{yi2025deep, mao2026packforcing}, and introduces only negligible overhead.

\paragraph{Why Frame Pruning Instead of Token Pruning} 
In our framework, the frame is the minimal unit for pruning, rather than individual tokens as in previous methods. This design choice is motivated by the observation that retaining a large number of tokens from multiple frames can introduce significant temporal gaps between early and recent frames in the RoPE-encoded key cache. Prior methods~\cite{yi2025deep} attempt to compact temporal positions by adjusting only the sink tokens while setting all other tokens to the same temporal position. However, this creates ambiguity in the 3D RoPE encoding: tokens from different frames but at the same spatial location are mapped to identical positional embeddings. We find that, compared to such fine-grained token pruning with position encoding errors, using coarser-grained frame pruning preserves accurate temporal encodings and yields better performance. Additional comparison experiments are provided in Appendix \ref{sec:more_exp}.

\subsection{Degradation-aware Training}
\label{sec:robust}

A key challenge in long-horizon autoregressive generation is that the historical context at inference time is not clean. It has been generated by the model itself and may contain accumulated errors, visual artifacts, or degraded representations. These errors tend to accumulate and amplify as the generated sequence grows. If the router is trained solely on clean context, it may over-trust historically early frames that have become unreliable due to error accumulation. Simulating such inference-time degraded context during training would require full long-context rollouts, incurring prohibitively high computational cost.

Inspired by Diffusion Forcing~\cite{chen2024diffusion}, we explicitly simulate context degradation during training. For a randomly selected subset $\mathcal{N}$ of historical frames, we apply random feature perturbations as the degradation operator:
\begin{equation}
\hat{f}_i = f_i + \beta \cdot \epsilon,
\label{eq:add_noise}
\end{equation}
where $\epsilon$ is random noise and $\beta$ controls the perturbation strength. The degraded context $\hat{\mathcal{C}}_{t-1}$ is constructed by replacing the selected frames with their degraded versions.
The perturbed historical Keys are used as inputs to the
Router, while both the perturbed Keys and Values are fed
into the Student's attention computation. The Teacher,
in contrast, uses the original, unperturbed historical KV cache to provide supervision.

To encourage the router to discount unreliable frames, we define a degradation-aware loss:
\begin{equation}
\mathcal{L}_{\mathrm{deg}}
=
\sum_{i\in\mathcal N}s_{i,\mathrm{retain}}^{\mathrm{noisy}}.
\label{eq:noise_loss}
\end{equation}
where $s_{i,\mathrm{retain}}^{\mathrm{noisy}}$ denotes the importance score predicted by the router for the degraded version of frame $i$. This loss teaches the router to distinguish between ``unreliable'' frames and ``genuinely valuable'' frames, improving its robustness to degraded context during inference.

The overall training objective combines all three losses:
\begin{equation}
\mathcal{L} = \mathcal{L}_{\mathrm{distill}} + \lambda_1 \mathcal{L}_{\mathrm{sparse}} + \lambda_2 \mathcal{L}_{\mathrm{deg}},
\label{eq:total_loss}
\end{equation}
where $\lambda_1$ controls the sparsity of context selection and $\lambda_2$ balances robustness.

\subsection{Dynamic Memory Management at Inference}
\label{sec:dynamic_memory}

During inference, the KV cache has a fixed budget, requiring a strategy to determine which frames to retain as generation progresses. Our dynamic memory management framework consists of three mechanisms: importance-based eviction, dynamic long-term memory initialization, and evolving.

\paragraph{Importance-based Frame Eviction}
Unlike FIFO eviction, which removes the oldest frames, our approach discards frames based on their predicted importance. Specifically, we preserve a set of sink frames $\mathcal{S}$ and recent frames $\mathcal{R}$, while treating the remaining frames as candidates $\mathcal{C}_{\mathrm{cand}}$, 
where $\mathcal{S}$, $\mathcal{R}$, and $\mathcal{C}_{\mathrm{cand}}$ are disjoint sets of latent frame indices, and $|\mathcal{S}|+|\mathcal{R}|+|\mathcal{C}_{\mathrm{cand}}|=N_h$, where $N_h$ denotes the total number of historical latent frames.
Importance scores are computed for frames in $\mathcal{C}_{\mathrm{cand}}$ using the router (Eq.~\ref{eq:router}), and frames with the lowest scores are evicted first. This strategy ensures that the most informative frames are retained under a fixed memory budget.

\paragraph{Dynamic Long-term Memory Initialization}
Previous methods often fix the initial frames as permanent sinks, motivated by observations in LLMs~\cite{xiao2023efficient} that attention tends to concentrate on early tokens. However, we observe that in autoregressive video generation, attention is frequently distributed across frames from diverse temporal positions (see Figure~\ref{fig:attention_weights}). 

Our analysis suggests that the effectiveness of sink frames is primarily attributed to their content quality and reliability, rather than their temporal order. Therefore, sink selection should be content-driven rather than position-dependent.
Instead of fixing the initial frames as sinks, we dynamically initialize the sink set. 
At the first eviction event, we select the top-$k$ frames (with $k = |\mathcal{S}|$) according to their importance scores as the initial sink set:
\begin{equation}
\mathcal{S}_0 = \mathrm{TopK}\bigl(\{s_i\}_{i \in \mathcal{C}_{t-1}}, \; k\bigr).
\label{eq:sink_init}
\end{equation}
This design ensures that sink frames correspond to the most informative historical content, rather than being biased toward early positions.

\paragraph{Long-term Memory Evolving}
Fixing the sink set after initialization can be suboptimal for long videos with significant motion or scene changes, as early long-term memory may become outdated. To enable the sink set to adapt to evolving content, we maintain an exponential moving average (EMA) of the importance scores for each frame:
\begin{equation}
\bar{s}_i^{(t)} = \alpha \, s_i^{(t)} + (1 - \alpha) \, \bar{s}_i^{(t-1)},
\label{eq:ema}
\end{equation}
where $s_i^{(t)}$ denotes the instantaneous importance score at step $t$, and $\alpha$ is the EMA coefficient. The EMA captures the sustained importance of a frame over time, smoothing out transient fluctuations. Smaller values of $\alpha$ place more emphasis on earlier, more reliable estimates, thereby mitigating error accumulation in later stages.

To further ensure robustness, a frame is considered eligible for long-term memory only if it persists for at least $N_p$ consecutive steps without being evicted. The sink set is then updated by selecting the top-$k$ frames from the union of existing sinks and eligible candidates:
\begin{equation}
\mathcal{S}_t = \mathrm{TopK}\bigl(\{\bar{s}_i^{(t)}\}_{i \in \mathcal{S}_{t-1} \cup \mathcal{P}_t}, \; k\bigr),
\label{eq:sink_update}
\end{equation}
where $\mathcal{P}_t$ denotes the set of eligible frames at step $t$.

This dynamic update mechanism allows the sink set to evolve with the video content: newly emerged high-quality frames can gradually replace outdated ones, enabling adaptive memory refinement while preserving long-term stability.

\paragraph{Why Dynamic Long-term Memory Matters}
The proposed long-term memory evolving mechanism addresses two complementary scenarios.
In \emph{stable scenes}, the sink set maintains long-term consistency by retaining frames that act as reliable anchors, thereby suppressing error accumulation and mitigating instability caused by local noise.
In \emph{dynamic scenes} with significant motion or scene transitions, a static sink set may impose outdated constraints, forcing the model to regress toward obsolete states. In contrast, the evolving mechanism allows newly generated frames to enter the long-term memory, enabling the context to adapt to the natural progression of the video.

\section{Experiments}
\label{sec:experiments}

\subsection{Experimental Settings}

\paragraph{Implementation Details}
We implement L2M on top of the Self Forcing~\cite{huang2025self} framework as our base model. The importance-aware frame router is realized as a 4-layer MLP with hidden dimension 768 and GeLU activations, taking as input the concatenation of the current chunk feature and each historical frame feature.
For inference, we set the KV cache budget $K$ to 21 latent frames, the sink set size $|\mathcal{S}|$ to 14 latent frames, the EMA coefficient $\alpha$ to 0.5, and the long-term memory update threshold $N_p$ to 6 steps.
The training objective weights are set to $\lambda_1 = 1.0$ and $\lambda_2 = 1.0$. To simulate degraded context, we apply feature perturbation to historical frames with a probability of 0.2 and $\beta=0.1$.
The router is jointly trained for 1000 steps with a batch size of 32 using the AdamW optimizer~\cite{loshchilov2017decoupled} with a learning rate of $1 \times 10^{-4}$. Training is conducted on 16 GPUs and takes approximately 6 hours. This setup balances training efficiency with effective learning of content-adaptive context management.

\paragraph{Evaluation Protocol}
Following standard practice~\cite{huang2025self, yi2025deep, cui2025self}, we evaluate on VBench-Long~\cite{huang2024vbench} using 128 prompts from MovieGen~\cite{polyak2024movie}.
Each prompt is refined using Qwen2.5-7B-Instruct~\cite{yang2025qwen3}. 
We report nine metrics from VBench-Long, including Aesthetic Quality, Background Consistency, Dynamic Degree, Imaging Quality, Motion Smoothness, Subject Consistency, Temporal Flickering, Quality Score, and Overall Consistency.
Overall Consistency is computed by averaging ViCLIP~\cite{wang2023internvid} scores over 5-second video segments. These metrics enable a comprehensive evaluation of the generated videos from multiple perspectives, including visual quality, temporal consistency, and semantic coherence.
We evaluate all methods on 30-second and 60-second videos, and additionally report 120-second results in Table~\ref{tab:main_results_120s}.

\subsection{Main Results}

\paragraph{Quantitative Comparison}
Table~\ref{tab:main_results} shows that L2M achieves the best overall trade-off between visual quality and long-term consistency on both 30-second and 60-second videos. 
It achieves the best results in aesthetic quality, background consistency, subject consistency, and overall
consistency, while maintaining competitive imaging quality and motion smoothness. These results demonstrate the effectiveness of dynamic memory management in preserving useful long-range context while maintaining natural motion.
Compared with Self Forcing, which uses FIFO-based memory eviction, our method better preserves visual fidelity and temporal coherence. Compared with Deep Forcing, which relies on a static sink and attention-score-based eviction, L2M further improves quality and consistency, showing the advantage of learnable importance prediction. Although Rolling Forcing is trained with sink tokens and yields relatively stable appearance, it suffers from limited dynamics, especially on longer videos. In contrast, L2M maintains stronger motion dynamics while avoiding obvious quality degradation.

\begin{table}[t]
\centering
\caption{%
\textbf{Quantitative comparison on long video generation.} 
We evaluate our method against open-source autoregressive video diffusion baselines on 30-second and 60-second videos using multiple quality metrics from VBench-Long~\cite{huang2024vbench}. All metrics follow the convention that higher values indicate better performance. 
The best results are in \textbf{bold} and the second best are \underline{underlined}.%
}\label{tab:main_results}
\resizebox{\textwidth}{!}{
\begin{tabular}{lccccccc|cc}
\toprule
Model & 
\makecell{\textbf{Aesthetic}\\\textbf{Quality}} & 
\makecell{\textbf{Background}\\\textbf{Consistency}} & 
\makecell{\textbf{Dynamic}\\\textbf{Degree}} & 
\makecell{\textbf{Imaging}\\\textbf{Quality}} & 
\makecell{\textbf{Motion}\\\textbf{Smoothness}} & 
\makecell{\textbf{Subject}\\\textbf{Consistency}} & 
\makecell{\textbf{Temporal}\\\textbf{Flickering}} & 
\makecell{\textbf{Quality}\\\textbf{Score}} &
\makecell{\textbf{Overall}\\\textbf{Consistency}} \\
\midrule
\rowcolor{black!5}
\multicolumn{10}{c}{\textit{30 seconds}} \\
CausVid & 61.80 & 90.70 & 54.69 & 66.98 & 98.06 & 87.69 & 96.80 & 79.18 & 22.92 \\
Rolling Forcing & 59.51 & 91.58 & 35.16 & \textbf{69.93} & \textbf{98.43} & \underline{89.23} & \underline{96.94} & 78.50 & 23.69 \\
Self Forcing & 60.28 & 90.81 & 44.53 & 68.73 & 97.97 & 87.08 & \textbf{97.00} & 78.39 & 24.10 \\
Deep Forcing & 61.17 & 91.90 & \textbf{59.38} & 68.08 & 97.70 & 87.56 & 96.12 & \underline{79.37} & \underline{24.55} \\
\rowcolor{black!2} \textbf{L2M} & \textbf{62.69} & \textbf{93.48} & \underline{57.81} & \underline{69.54} & \underline{98.17} & \textbf{90.99} & 96.71 & \textbf{81.15} & \textbf{24.78} \\
\midrule
\rowcolor{black!5}
\multicolumn{10}{c}{\textit{60 seconds}} \\
CausVid & 61.43 & 89.41 & \underline{54.69} & 66.57 & 98.12 & 85.48 & \underline{96.96} & 78.49 & 22.30 \\
Rolling Forcing & 58.57 & 90.52 & 35.94 & \textbf{69.88} & \textbf{98.36} & 86.74 & 96.85 & 77.66 & 23.01 \\
Self Forcing & 57.01 & 87.52 & 33.59 & 67.08 & 98.06 & 81.22 & \textbf{97.33} & 75.23 & 21.58 \\
Deep Forcing & \underline{59.74} & \underline{92.84} & 53.12 & 67.20 & 97.66 & \underline{89.91} & 96.21 & \underline{79.17} & \underline{24.24} \\
\rowcolor{black!2} \textbf{L2M} & \textbf{62.17} & \textbf{93.20} & \textbf{56.25} & \underline{69.43} & \underline{98.15} & \textbf{90.29} & 96.70 & \textbf{80.73} & \textbf{24.54} \\
\bottomrule
\end{tabular}
}
\end{table}

\begin{figure}
    \centering
    \includegraphics[width=0.95\linewidth]{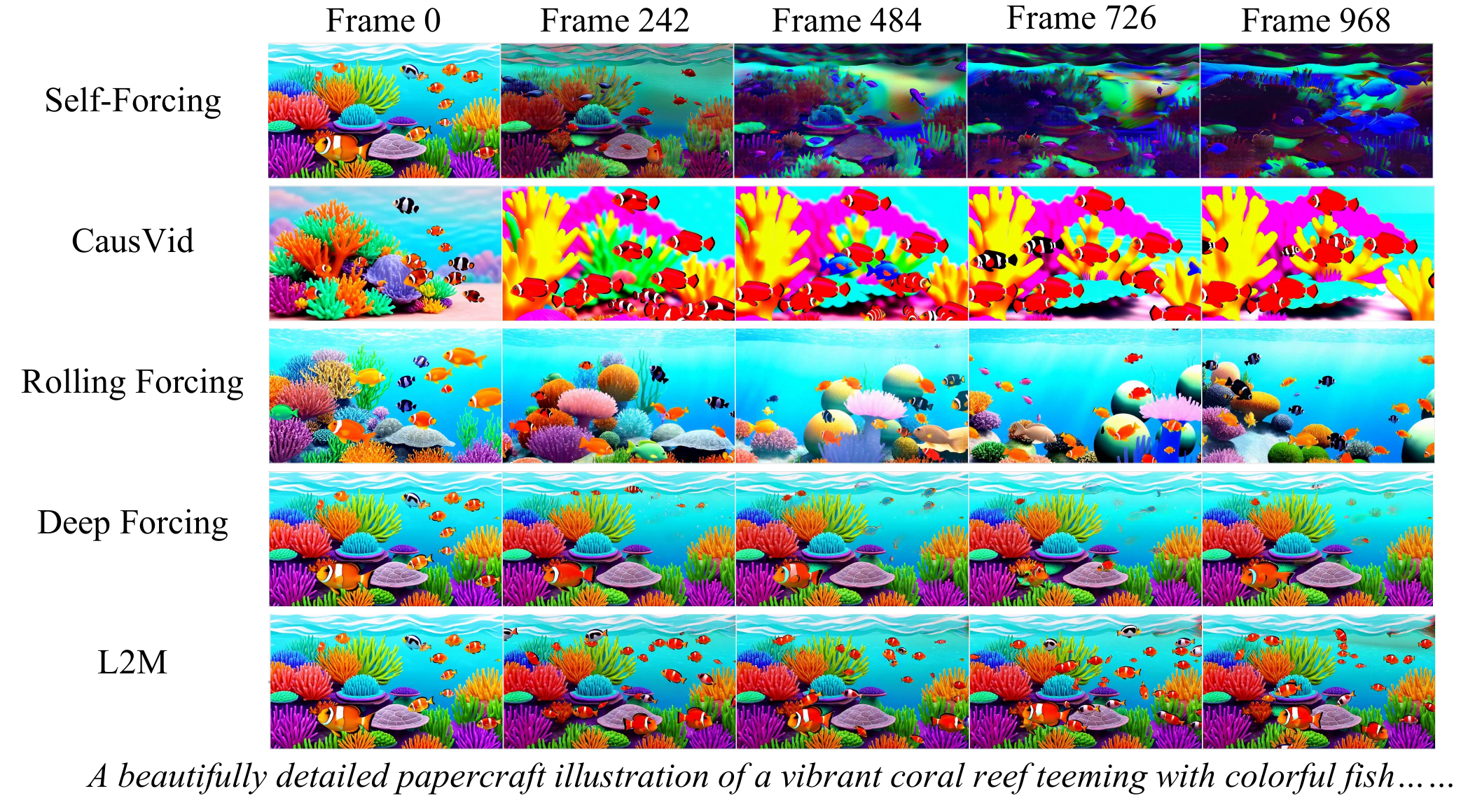}
   \caption{Qualitative comparison with baseline methods. Our method demonstrates superior visual quality and stronger overall consistency.}
    \label{fig:qualitative}
    \vspace{-15pt}
\end{figure}

\paragraph{Qualitative Results}
Figure~\ref{fig:qualitative} presents qualitative comparisons with existing methods. Self Forcing, which employs FIFO-based memory eviction, exhibits noticeable color drift and texture degradation in later frames. CausVid also shows oversaturation and visual distortion in later stages. Rolling Forcing, trained with attention sinks, produces relatively stable results; however, it exhibits weaker background consistency. Deep Forcing uses a static long-term memory and evicts frames based on attention scores, maintaining better visual quality but still exhibits identity inconsistency and gradual image quality degradation over time.
In contrast, L2M predicts learnable importance scores and dynamically manages memory, maintaining consistent visual quality throughout generation while preserving natural motion dynamics.

\begin{wrapfigure}{tr}{0.47\textwidth}
    \centering
    \vspace{-15pt}
    \includegraphics[width=0.47\textwidth]{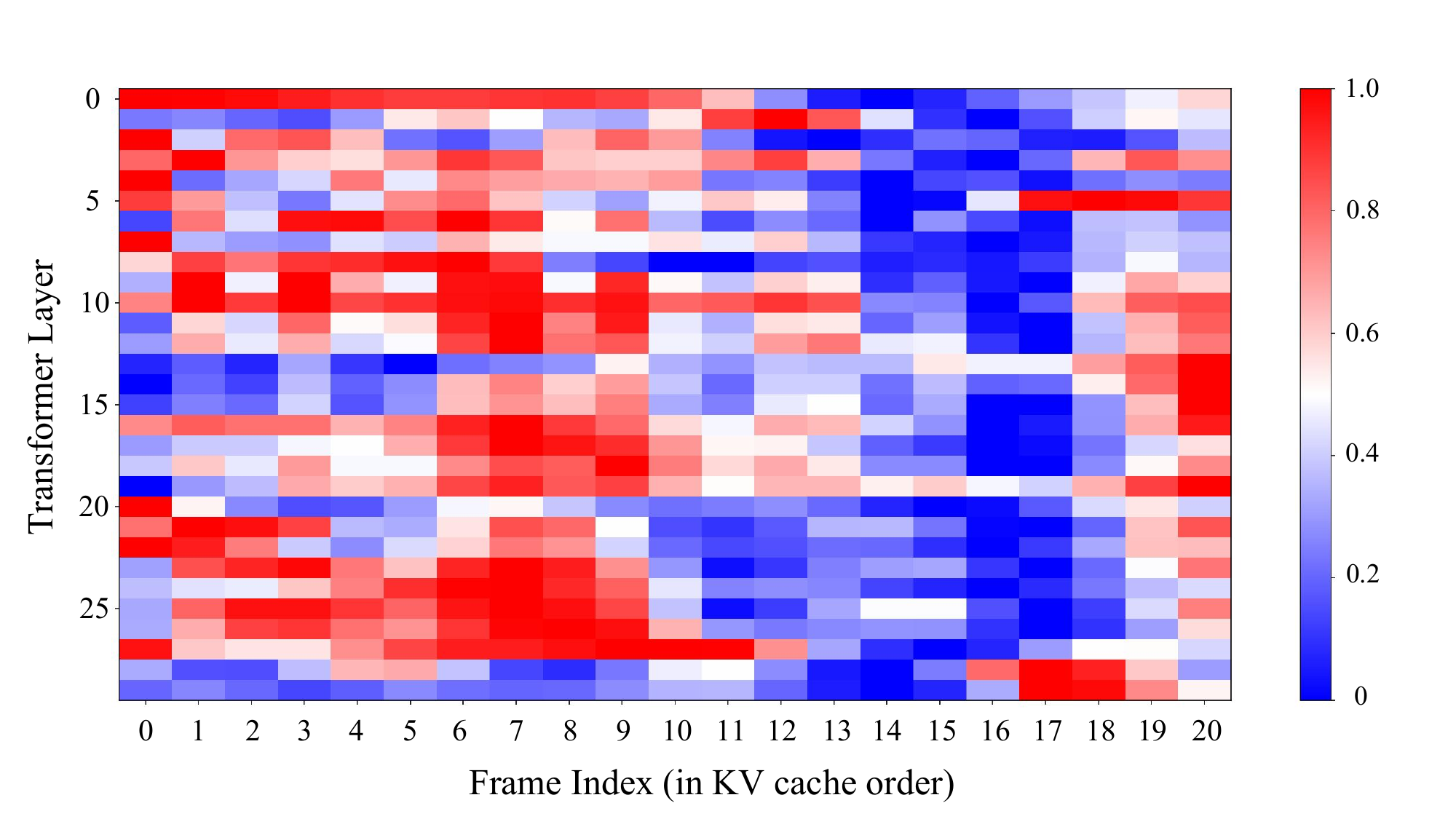}
    \vspace{-12pt}
    \caption{Visualization of layer-wise min-max normalized predicted importance scores.}
    \label{fig:norm_router_prediction_pattern}
\end{wrapfigure}

\paragraph{Router Prediction Patterns}
We visualize the router prediction patterns in Figure~\ref{fig:norm_router_prediction_pattern}. We observe structured patterns across frame chunks: early high-quality frames and their neighboring chunks tend to receive higher importance scores, while intermediate frames are often assigned low scores and appear redundant. 

This partly explains why prior methods that retain initial and nearby frames while discarding middle ones can achieve reasonable performance. However, not all early or temporally adjacent frames are equally important. The importance distribution varies across layers, and some such frames still receive low scores, indicating they should not be persistently preserved. This highlights the limitation of static heuristics and motivates our dynamic eviction and evolving memory design, which selectively retains truly informative frames based on learned importance.

\subsection{Ablation Studies}
\label{sec:ablation}
We conduct comprehensive ablation studies to evaluate the contribution of each component. More ablation results are provided in Appendix \ref{sec:more_exp}.

\paragraph{Effect of Individual Components}
Table~\ref{tab:ablation} shows the contribution of each component on 30-second videos. Starting from the base model, introducing the router yields clear gains in both Quality Score and Overall Consistency, indicating that, compared to the naive FIFO strategy, routing can better select informative context for generation. Incorporating degradation-aware training further improves performance, suggesting enhanced robustness to long-sequence content degradation. Enabling dynamic sink initialization brings additional gains, demonstrating that long-term memory selection should be content-aware rather than position-based. Finally, allowing the dynamic sink to evolve during generation achieves the best performance, confirming that adaptive memory updating is critical for maintaining visual quality while preserving long-term consistency.

\begin{table}[h]
  \caption{\textbf{Ablation study on individual components of our method.} Results are evaluated on 30-second videos.}
  \label{tab:ablation}
  \centering
  \small
  \begin{tabular}{@{}cccccc@{}}
    \toprule
    Router & Degradation & Dynamic Sink Init. & Dynamic Sink Evolving & Quality Score$\uparrow$  & Overall Cons.$\uparrow$ \\
    \midrule
                &               &       &         & 78.39 & 24.10 \\
    \checkmark  &               &        &       & 80.32 & 24.44 \\
    \checkmark  & \checkmark    &            &   & 80.56 & 24.52 \\
    \checkmark  & \checkmark    & \checkmark  &  & 80.88 & 
    24.70 \\
    \checkmark  & \checkmark    & \checkmark   & \checkmark  & \textbf{81.15} & 
    \textbf{24.78} \\
    \bottomrule
  \end{tabular}
\end{table}

\paragraph{Comparison with Alternative Importance Estimation}
To validate the effectiveness of our importance prediction router, we compare it with two baselines: random scoring and attention-based scoring. Random selection achieves a Quality Score of 79.92 and an Overall Consistency of 24.43, indicating that unstructured memory management fails to preserve useful context in long video generation. Attention-based scoring improves to 80.68 and 24.55, suggesting that attention weights provide a reasonable proxy for frame importance. 
In contrast, our learned importance predictor achieves the best performance, reaching 81.15 in Quality Score and 24.78 in Overall Consistency, demonstrating the effectiveness of learning frame importance over heuristic or random strategies.

\begin{figure}
    \centering
    \includegraphics[width=1.0\linewidth]{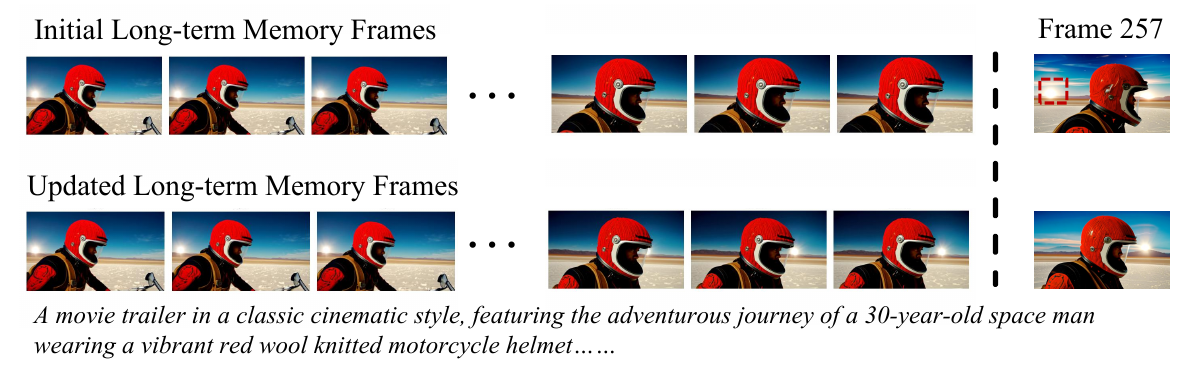}
    \caption{Qualitative analysis of dynamic long-term memory evolution. After memory updates, newly emerged informative frames are incorporated into sink memory, enabling better motion tracking and reducing long-term generation artifacts.}
    \label{fig:memory_update}
\end{figure}

\paragraph{Qualitative Analysis of Dynamic Memory}
We further provide a qualitative ablation study in Figure~\ref{fig:memory_update} to analyze the effect of dynamic long-term memory. We observe that incorporating dynamic sink memory evolving consistently improves generation quality. This improvement mainly stems from introducing more content-relevant, rather than position-biased, information into long-term memory, while continuously updating the memory to adapt to evolving scenes.

As shown in Figure~\ref{fig:memory_update}, taking the first-layer long-term memory as an example, we observe that after memory updates, frames capturing the positional changes of the sun are incorporated into the sink memory. This enables the model to better track the motion of the sun, avoiding temporal regression artifacts and duplicated suns that appear in the variant without long-term memory updates.

\paragraph{Quantifying Error Accumulation.}
To quantify temporal error accumulation, we report ViCLIP scores for consecutive 5-second segments over the 30-second generation horizon, tracking the evolution of text-video consistency.
As shown in Table~\ref{tab:viclip_segments}, all methods exhibit lower ViCLIP scores in the final segment than in the initial one, but with different degradation rates. Self Forcing, Deep Forcing, and L2M start at the same score of $25.28$, yet diverge as generation proceeds. Self Forcing suffers the largest decline among these three methods, dropping by $3.49$ points, while CausVid, Rolling Forcing, and Deep Forcing decline by $2.52$, $2.69$, and $1.08$ points, respectively. In contrast, L2M maintains a nearly stable score after the first 5--10\,s segment, decreasing from $24.79$ to $24.54$ in the final segment, with the smallest overall decline of $0.74$. L2M also achieves the highest segment-wise scores from $10$\,s onward and the best overall score of $24.78$. These results highlight the benefit of learned context selection and EMA-based memory updates in preserving informative historical frames and mitigating the long-horizon degradation caused by static or FIFO-based
context management.

\begin{table}[t]
  \centering
  \caption{ViCLIP score on consecutive 5-second segments of 30-second videos. Overall Consistency is the average of the six segment scores. $\Delta$ is the drop from the first segment to the last. Higher is better for segment and overall scores; lower is better for $\Delta$. Best results are in \textbf{bold}.}
  \label{tab:viclip_segments}
  \small
  \begin{tabular}{lcccccccc}
    \toprule
    Method & 0--5\,s & 5--10\,s & 10--15\,s & 15--20\,s & 20--25\,s & 25--30\,s & $\Delta \downarrow$ & Overall $\uparrow$ \\
    \midrule
    CausVid         & 24.62          & 23.27          & 22.87          & 22.35          & 22.32          & 22.10          & 2.52          & 22.92 \\
    Rolling Forcing & 25.19          & 24.51          & 23.92          & 23.08          & 22.91          & 22.50          & 2.69          & 23.69 \\
    Self Forcing    & {25.28} & \textbf{25.28} & 24.70          & 24.52          & 23.11          & 21.79          & 3.49          & 24.11 \\
    Deep Forcing    & {25.28} & 24.74          & 24.47          & 24.39          & 24.24          & 24.20          & 1.08          & 24.55 \\
    L2M             & \textbf{25.28} & 24.79          & \textbf{24.72} & \textbf{24.71} & \textbf{24.65} & \textbf{24.54} & \textbf{0.74} & \textbf{24.78} \\
    \bottomrule
  \end{tabular}
\end{table}

\section{Conclusion}
\label{sec:conclusion}
We proposed \textbf{L2M}, a learned context management framework for long autoregressive video generation. By dynamically selecting and updating historical frames based on their predicted importance, L2M overcomes the limitations of static caches and fixed sinks. Through importance-aware prediction, degradation-aware training, and dynamic memory updating, L2M better utilizes a fixed KV-cache budget to improve long-range consistency and visual fidelity. Extensive experiments demonstrate its advantages over existing baselines.

\bibliographystyle{plainnat}
\bibliography{references}

\appendix
\clearpage

\section{Broader Impact}
\label{sec:broad_impact}
This work improves long-horizon autoregressive video generation by introducing a learned dynamic memory management mechanism. On the positive side, it can benefit a wide range of applications such as video editing, content creation, simulation, and storytelling by enabling more temporally consistent and visually coherent long video synthesis under limited memory budget. 

On the potential negative side, as with most generative video models pretrained on large-scale internet data, our method may inherit and amplify existing biases and stereotypes present in the training distribution \cite{cho2023dall, naik2023social, ross2020measuring}. Such biases may manifest in generated visual content, potentially affecting fairness and representation in downstream applications.

In addition, improved long-horizon video generation capabilities may also increase the risk of misuse, including the creation of highly realistic DeepFakes \cite{lyu2020deepfake} and other synthetic media that could be used for misinformation or impersonation. While our method reduces the computational and memory cost of long video generation, lowering the barrier to content creation, it also emphasizes the importance of responsible deployment.

We therefore encourage the development of appropriate governance frameworks, usage guidelines, and detection mechanisms to ensure that generative models are used in a safe and responsible manner, and to mitigate potential negative societal impacts.

\section{Limitations and Future Work}
\label{sec:limitation}
While L2M substantially improves context management for autoregressive video generation, several limitations remain.
First, the Router introduces additional computational overhead during both training and inference, although this cost is largely mitigated by its lightweight design. Second, the current framework operates at the frame level; extending it to finer-grained, token-level management could further improve cache efficiency and flexibility. Third, the degradation-aware training relies on hand-crafted degradation operators. Learning the degradation distribution directly from real generation errors may lead to more robust and adaptive behavior.
Future work may explore hierarchical memory architectures with multiple levels of compression, as well as extensions to interactive generation settings, where user inputs dynamically influence context management decisions.

\section{More Implementation Details}
\subsection{Attention with Masking}
\label{sec:supp_attn_with_mask}
To enable end-to-end optimization, the sampling score $s_i$ in Eq.~\ref{eq:router} are converted into a binary mask using the Gumbel-Softmax trick. This operator yields one-hot outputs while remaining differentiable, thereby allowing gradient-based training.

In addition, when token pruning is applied within a mini-batch, the learned masks are typically irregular, resulting in different samples retaining varying numbers of tokens. Physically removing the pruned tokens would therefore lead to variable sequence lengths across samples, which is incompatible with efficient batch computation. 

Simply zeroing out pruned tokens is also insufficient to suppress their influence, as they can still affect other tokens through the softmax normalization in self-attention:
\begin{equation}
    \mathbf{A} = \operatorname{softmax}\left(
    \frac{\mathbf{Q}\mathbf{K}^T}{\sqrt{C}}
    \right).
\end{equation}

To fully eliminate such influence, following the strategy in~\cite{rao2021dynamicvit}, attention masking is employed to compute the masked attention matrix:
\begin{align}
    &\mathbf{P} = \mathbf{Q}\mathbf{K}^T/\sqrt{C} \in \mathbb{R}^{N_q\times N_k},\\
    &\mathbf{G}_{ij} = 
    \begin{cases}
    1,& i=j,\\
    \hat{\mathbf{D}}_j,& i\neq j,
    \end{cases}\quad 1\le i\le N_q,\ 1\le j\le N_k,\\
    &\tilde{\mathbf{A}}_{ij} = \frac{\exp(\mathbf{P}_{ij})\mathbf{G}_{ij}}{\sum_{k=1}^{N_k}\exp(\mathbf{P}_{ik})\mathbf{G}_{ik}},\quad 1\le i\le N_q,\ 1\le j\le N_k.
    \label{eq:attn_mask}
\end{align}
This construction can be interpreted as a graph in which $\mathbf{G}_{ij}=1$ indicates that token $j$ is allowed to contribute to the update of token $i$. Self-loops are added for numerical stability. Importantly, when $\hat{\mathbf{D}}_j=0$, token $j$ only preserves its self-loop and does not affect any other token. As a result, $\tilde{\mathbf{A}}$ is equivalent to performing attention over the retained tokens only, while keeping a constant $N_q\times N_k$ shape during training.

\subsection{Pseudo Algorithm}
We provide the pseudo algorithm of our method in Algorithm~\ref{alg:l2m}.
\begin{algorithm}[!htbp]
\caption{L2M: Learned Context Management for Long-Horizon Autoregressive Video Generation}
\label{alg:l2m}
\begin{algorithmic}[1]
\Require Router $R$, teacher $\phi_t$, student $\phi_s$, total latent frames $T$, frames per chunk $B_f$, cache budget $K$, sink budget $K_s$, recent budget $K_r$, patience $N_p$, EMA coefficient $\alpha$, sparsity weight $\lambda_1$, degradation weight $\lambda_2$
\State $N_c \leftarrow \lceil T/B_f \rceil$
\State Initialize memory $\mathcal{C}_0 \leftarrow \emptyset$, sink set $\mathcal{S} \leftarrow \emptyset$, recent set $\mathcal{R} \leftarrow \emptyset$, and EMA scores $\bar{s}_i^{(0)} \leftarrow 0$
\For{$b=1$ to $N_c$}
    \State $n_b \leftarrow \min((b-1)B_f, T)$
    \State $B_b \leftarrow \min(B_f, T-(b-1)B_f)$
    \State $z_b \leftarrow [x_{(b-1)B_f+1}, \ldots, x_{(b-1)B_f+B_b}]$
    \State Extract current chunk feature $\bar{f}_b$ and historical frame features $\{\bar{f}_i\}_{i=1}^{n_b}$
    \If{$n_b > 0$}
        \If{training}
            \State Sample degraded subset $\mathcal{N} \subseteq \{1,\ldots,n_b\}$
            \For{$i=1$ to $n_b$}
                \If{$i \in \mathcal{N}$}
                    \State $\hat{f}_i \leftarrow \bar{f}_i + \beta \epsilon$
                    \Comment{\textcolor{gray}{Eq.~(\ref{eq:add_noise})}}
                    \State $s_i \leftarrow R(\bar{f}_b, \hat{f}_i)$
                \Else
                    \State $s_i \leftarrow R(\bar{f}_b, \bar{f}_i)$
                \EndIf
                \Comment{\textcolor{gray}{Eq.~(\ref{eq:router})}}
                \State $m_i \leftarrow \mathrm{GumbelSoftmax}(s_i)$
                \Comment{\textcolor{gray}{Frame-level selection mask}}
            \EndFor
            \State $\tilde{\mathcal{C}}_{b-1} \leftarrow
            \{(\mathbf{K}_i^l,\mathbf{V}_i^l)
            \mid m_{i,\mathrm{retain}}=1\}$
            \State Adjust Temporal RoPE of the historical key cache
            \Comment{\textcolor{gray}{Eq.~(\ref{eq:time_adjust})}}
            \State $y^* \leftarrow \phi_t(z_b, \mathcal{C}_{b-1})$
            \State $y \leftarrow \phi_s(z_b, \tilde{\mathcal{C}}_{b-1})$
            \Comment{\textcolor{gray}{Use Eq.~(\ref{eq:attn_mask}) for parallel, differentiable forward passes}}
            \State $\mathcal{L}_{\mathrm{distill}} \leftarrow \|y-y^*\|_2^2$
            \Comment{\textcolor{gray}{Eq.~(\ref{eq:distill_loss})}}
            \State $\mathcal{L}_{\mathrm{sparse}} \leftarrow
            \frac{1}{n_b}\sum_{i=1}^{n_b}m_{i,\mathrm{retain}}$
            \Comment{\textcolor{gray}{Eq.~(\ref{eq:sparse_loss})}}
            \State $\mathcal{L}_{\mathrm{deg}} \leftarrow
            \sum_{i\in\mathcal{N}}s_{i,\mathrm{retain}}^{\mathrm{noisy}}$
            \Comment{\textcolor{gray}{Eq.~(\ref{eq:noise_loss})}}
            \State $\mathcal{L} \leftarrow
            \mathcal{L}_{\mathrm{distill}}
            +\lambda_1\mathcal{L}_{\mathrm{sparse}}
            +\lambda_2\mathcal{L}_{\mathrm{deg}}$
            \Comment{\textcolor{gray}{Eq.~(\ref{eq:total_loss})}}
            \State Update router $R$ by minimizing $\mathcal{L}$
        \Else
            \State Compute frame importance scores
            $\{s_i^{(b)}\}_{i=1}^{n_b}
            \leftarrow R(\bar{f}_b,\{\bar{f}_i\}_{i=1}^{n_b})$
            \If{$\mathcal{S}=\emptyset$}
                \State $\mathcal{S} \leftarrow
                \mathrm{TopK}(\{s_i^{(b)}\}_{i=1}^{n_b},K_s)$
                \Comment{\textcolor{gray}{Eq.~(\ref{eq:sink_init})}}
            \EndIf
            \State Update EMA scores
            $\bar{s}_i^{(b)} \leftarrow
            \alpha s_i^{(b)}+(1-\alpha)\bar{s}_i^{(b-1)}$
            \Comment{\textcolor{gray}{Eq.~(\ref{eq:ema})}}
            \State $\mathcal{P}_b \leftarrow
            \{i \mid i \text{ survives for } N_p
            \text{ consecutive steps}\}$
            \State $\mathcal{S} \leftarrow
            \mathrm{TopK}\bigl(
            \{\bar{s}_i^{(b)}\}_{i\in\mathcal{S}\cup\mathcal{P}_b},
            K_s\bigr)$
            \Comment{\textcolor{gray}{Eq.~(\ref{eq:sink_update})}}
            \State $\mathcal{C}_{\mathrm{cand}} \leftarrow
            \mathcal{C}_{b-1}\setminus(\mathcal{S}\cup\mathcal{R})$
            \State Evict the lowest-scoring frames in
            $\mathcal{C}_{\mathrm{cand}}$ until
            $|\mathcal{S}|+|\mathcal{R}|+
            |\mathcal{C}_{\mathrm{cand}}|\leq K-B_b$
            \Comment{\textcolor{gray}{Make room for the incoming chunk}}
            \State Adjust Temporal RoPE of the historical key cache
            \Comment{\textcolor{gray}{Eq.~(\ref{eq:time_adjust})}}
            \State Generate the current chunk $z_b$ using the managed context
            $(\mathcal{S},\mathcal{R},\mathcal{C}_{\mathrm{cand}})$
            \State Update memory $\mathcal{C}_b$ with the newly generated chunk
            \State Update recent set $\mathcal{R}$ with the most recent $K_r$ frames in $\mathcal{C}_b$
        \EndIf
    \Else
        \State Generate the first chunk $z_1$ without historical context
        \State Initialize memory $\mathcal{C}_1$ with the newly generated chunk
    \EndIf
\EndFor
\end{algorithmic}
\end{algorithm}

\section{Differences from Existing Work}
Although recent autoregressive video generation methods have substantially improved long-horizon synthesis, they mostly differ from our approach in that they rely on \emph{static} context management policies. For example, CausVid~\cite{yin2025slow} and Self Forcing~\cite{huang2025self} primarily focus on bridging the train-inference gap and improving robustness in causal generation, but they still manage historical context using hand-crafted cache policies rather than learning which frames are truly informative. As a result, their retained context is determined by predefined rules instead of content-aware importance, which may preserve redundant frames while discarding useful ones.

Rolling Forcing~\cite{liu2025rolling} extends autoregressive generation to longer sequences by introducing rolling memory and fixed sink-style context retention, which improves temporal stability but remains heuristic in nature. Similarly, Deep Forcing~\cite{yi2025deep} augments the memory with deeper long-term tokens, and Pack Forcing~\cite{mao2026packforcing} compresses context through packing strategies to reduce cache usage. However, both methods still depend on static or heuristic criteria for deciding what to keep, and thus cannot adapt their memory organization to different scenes, motions, or video stages. In contrast, our method learns an explicit importance router that dynamically ranks frames based on their predicted contribution to future generation.

More broadly, unlike prior sink-based or FIFO-based approaches that either permanently preserve early frames or evict frames according to recency, L2M introduces a learned and evolving memory mechanism. It dynamically initializes the sink set from content-relevant frames and continuously updates long-term memory using predicted importance scores, enabling the model to retain informative historical context while discarding redundant or outdated frames. This makes L2M fundamentally different from existing long-video generation methods, which either rely on fixed memory heuristics or use compression rules that are not content-adaptive.

\section{More Experiments}
\label{sec:more_exp}
\paragraph{Longer Video Generation Extrapolation}
We further evaluate L2M on 120-second video generation, corresponding to a $24\times$ extrapolation beyond the 5-second training horizon. As shown in Table~\ref{tab:main_results_120s}, L2M maintains strong performance despite the substantially extended generation duration, achieving the highest Quality Score ($80.30$) and Overall Consistency ($24.47$). It also obtains the best results in Aesthetic Quality, Dynamic Degree, Imaging Quality, and Subject Consistency. These results demonstrate that the proposed dynamic context management strategy generalizes effectively to generation horizons far beyond those encountered during training.

\begin{table}[t]
\centering
\caption{%
\textbf{Quantitative comparison on 120-second long video generation.} 
We evaluate our method against open-source autoregressive video diffusion baselines using multiple quality metrics from VBench-Long~\cite{huang2024vbench}. All metrics follow the convention that higher values indicate better performance. 
The best results are in \textbf{bold} and the second best are \underline{underlined}.%
}\label{tab:main_results_120s}
\resizebox{\textwidth}{!}{
\begin{tabular}{lccccccc|cc}
\toprule
Model & 
\makecell{\textbf{Aesthetic}\\\textbf{Quality}} & 
\makecell{\textbf{Background}\\\textbf{Consistency}} & 
\makecell{\textbf{Dynamic}\\\textbf{Degree}} & 
\makecell{\textbf{Imaging}\\\textbf{Quality}} & 
\makecell{\textbf{Motion}\\\textbf{Smoothness}} & 
\makecell{\textbf{Subject}\\\textbf{Consistency}} & 
\makecell{\textbf{Temporal}\\\textbf{Flickering}} & 
\makecell{\textbf{Quality}\\\textbf{Score}} &
\makecell{\textbf{Overall}\\\textbf{Consistency}} \\
\midrule
CausVid & 61.13 & 88.46 & 45.31 & 66.20 & 98.15 & 83.73 & 97.08 & 77.22 & 21.60 \\
Rolling Forcing & 57.86 & 89.87 & 36.72 & \underline{69.47} & \textbf{98.31} & 85.03 & 96.80 & 77.06 & 22.59 \\
Self Forcing & 51.14 & 84.00 & 28.91 & 63.09 & 98.10 & 74.33 & \textbf{97.52} & 71.48 & 17.04 \\
Deep Forcing & \underline{61.48} & \textbf{93.06} & 44.53 & 68.88 & 98.13 & \underline{89.56} & \underline{97.00} & \underline{79.59} & \underline{24.33} \\
\rowcolor{black!2} \textbf{L2M} & \textbf{61.88} & \underline{92.93} & \textbf{53.12} & \textbf{69.56} & \underline{98.14} & \textbf{89.72} & 96.71 & \textbf{80.30} & \textbf{24.47} \\
\bottomrule
\end{tabular}
}
\end{table}

We further extend the evaluation horizon to 240 seconds. Notably, the advantage of L2M over its Self Forcing backbone becomes progressively larger as the generation horizon increases. The improvements in Quality Score grow from $2.76$ at 30 seconds to $5.50$, $8.82$, and $12.35$ at 60, 120, and 240 seconds, respectively. Similarly, the gains in Overall Consistency increase from $0.68$ to $2.96$, $7.43$, and $10.54$. At 240 seconds, L2M maintains a Quality Score of $80.16$ and an Overall Consistency of $24.30$, whereas Self Forcing degrades to $67.81$ and $13.76$, respectively. L2M also outperforms the strongest competing method at this horizon, LongLive, which obtains $79.43$ and $24.03$ on the two metrics. These results provide direct evidence that adaptive context management becomes increasingly important over extended generation horizons: under a fixed KV-cache budget, the model must preserve informative context while preventing accumulated errors and outdated information from propagating to subsequent frames.

\begin{table}[t]
  \centering
  \caption{\textbf{Quantitative comparison on 240-second long video generation.} Higher is better for all metrics. The best results are in \textbf{bold}, and the second-best results are \underline{underlined}.}
  \label{tab:vbench240}
  \resizebox{\textwidth}{!}{%
  \begin{tabular}{lccccccc|cc}
    \toprule
    Model & Aesthetic & Background & Dynamic & Imaging & Motion & Subject & Temporal & Quality & Overall \\
          & Quality & Consistency & Degree & Quality & Smoothness & Consistency & Flickering & Score & Consistency \\
    \midrule
    CausVid         & 60.67             & 87.78             & \underline{44.53} & 66.00             & 98.21             & 82.54             & 97.24             & 76.80             & 21.12 \\
    Rolling Forcing & 57.40             & 89.48             & 36.72             & 69.36             & \underline{98.27} & 84.11             & 96.76             & 76.68             & 22.40 \\
    Self Forcing    & 44.89             & 81.27             & 24.22             & 58.75             & 97.92             & 68.82             & \underline{97.46} & 67.81             & 13.76 \\
    Deep Forcing    & 59.70             & 89.80             & 42.19             & 66.84             & 97.68             & 83.82             & 96.40             & 76.62             & 23.33 \\
    LongLive        & \underline{61.23} & \underline{92.39} & 40.62             & \underline{69.38} & \textbf{98.31}    & \underline{89.04} & \textbf{97.58}    & \underline{79.43} & \underline{24.03} \\
    L2M             & \textbf{61.61}    & \textbf{92.74}    & \textbf{52.34}    & \textbf{69.60}    & 98.16             & \textbf{89.48}    & 96.77             & \textbf{80.16}    & \textbf{24.30} \\
    \bottomrule
  \end{tabular}%
  }
\end{table}

\paragraph{Human Evaluation}
We further conduct a two-alternative forced-choice (2AFC) user study comparing L2M with CausVid, Rolling Forcing, Deep Forcing, and Self Forcing. For each pair of videos, participants indicate their preference according to four criteria: (1) scene consistency throughout the video; (2) the naturalness and dynamics of subject and camera motion; (3) consistency in the identity and appearance of the main subject; and (4) overall realism and visual quality. Ten participants each evaluate 30 pairs generated from randomly sampled MovieGen prompts, with no prompt repeated within a participant's session, resulting in 300 pairwise comparisons in total.

Table~\ref{tab:human} reports the percentage of votes favoring L2M over each baseline. L2M is preferred over every competing method across all four criteria, with preference rates ranging from $76.2\%$ to $100.0\%$. In particular, when compared with CausVid, Deep Forcing, and Self Forcing, L2M receives more than $90\%$ of the votes for scene consistency, subject consistency, and overall quality. These results corroborate the quantitative evaluation and demonstrate that the improvements of L2M are also clearly perceptible to human observers.

\begin{table}[t]
  \centering
  \caption{Human evaluation using two-alternative forced choice (2AFC). Each entry denotes the percentage of votes favoring L2M over the corresponding baseline. Higher is better. The best results are in \textbf{bold}, and the second-best results are \underline{underlined}.}
  \label{tab:human}
  \small
  \begin{tabular}{lcccc}
    \toprule
    Baseline & Scene & Motion & Subject & Overall \\
             & Consistency & Quality & Consistency & Quality \\
    \midrule
    CausVid         & 90.7             & \textbf{100.0}    & 90.7             & \textbf{96.0}    \\
    Rolling Forcing & 80.8             & 76.2              & 84.5             & 78.8             \\
    Self Forcing    & \underline{93.0} & \underline{97.6} & \underline{93.0} & 93.0             \\
    Deep Forcing    & \textbf{94.0}    & 85.0              & \textbf{96.0}    & \underline{94.0} \\
    \bottomrule
  \end{tabular}
\end{table}

\paragraph{Comparison on Additional Benchmark}
To assess whether the benefits of L2M generalize beyond the MovieGen prompt distribution used in VBench-Long, we further evaluate all methods on LoCoT2V-Bench~\cite{zheng2025locot2v}. This benchmark features long-form, multi-scene prompts with substantially greater linguistic and structural complexity. Following our main evaluation setting, we generate 30-second videos and evaluate them using the official Perceptual Quality (PQ) metric, which measures frame-level visual fidelity.

As shown in Table~\ref{tab:locot2v}, L2M achieves the highest PQ score of $77.37$, outperforming the strongest baseline, Rolling Forcing, by $1.84$ points. It also surpasses Deep Forcing, CausVid, and Self Forcing by $3.56$, $6.61$, and $6.93$ points, respectively. These results demonstrate that L2M remains effective at preserving visual quality under a distinct and more challenging prompt distribution, supporting its generalization beyond VBench-Long. 

\begin{table}[t]
  \centering
  \caption{\textbf{Quantitative comparison on LoCoT2V-Bench.}
  We report Perceptual Quality (PQ) for 30-second videos. Higher is better. The best result is in \textbf{bold}, and the second-best result is \underline{underlined}.}
  \label{tab:locot2v}
  \small
  \begin{tabular}{lc}
    \toprule
    Model & \textbf{Perceptual Quality} $\uparrow$ \\
    \midrule
    CausVid         & 70.76 \\
    Rolling Forcing & \underline{75.53} \\
    Self Forcing    & 70.44 \\
    Deep Forcing    & 73.81 \\
    \rowcolor{black!2}
    \textbf{L2M}    & \textbf{77.37} \\
    \bottomrule
  \end{tabular}
\end{table}

\paragraph{Compatibility with Additional Baseline}
To investigate whether L2M transfers to different baseline models, we integrate it into LongLive by replacing its original fixed sink-and-recent-window policy with our learned context selection mechanism, while keeping the underlying video generator unchanged. As shown in Table~\ref{tab:longlive_transfer}, L2M improves the Quality Score from $80.73$ to $81.20$ and Overall Consistency from $24.58$ to $24.84$. These results demonstrate that L2M can be readily incorporated into other autoregressive long-video generation frameworks without retraining their underlying generators.

\begin{table}[t]
  \centering
  \caption{\textbf{Integration of L2M into LongLive.}
  Results are evaluated on 30-second videos using 128 MovieGen prompts. Higher is better for both metrics.}
  \label{tab:longlive_transfer}
  \small
  \begin{tabular}{lcc}
    \toprule
    Method &
    \textbf{Quality Score} $\uparrow$ &
    \textbf{Overall Consistency} $\uparrow$ \\
    \midrule
    LongLive       & 80.73          & 24.58          \\
    \rowcolor{black!2}
    LongLive + L2M & \textbf{81.20} & \textbf{24.84} \\
    \bottomrule
  \end{tabular}
\end{table}

\paragraph{Inference Complexity Analysis}
Our method introduces an additional Router for importance prediction. We analyze its overhead in terms of both computation and memory. Specifically, we evaluate the inference cost for generating 30-second videos on an A100 GPU. Results are averaged over 10 samples.
The results show that the Router incurs only negligible computational overhead. This efficiency mainly stems from two factors: (1) the Router is lightweight, and (2) it is executed only at the first timestep rather than at every timestep. In addition, our method requires tracking the lifespan of context frames and maintaining an EMA score for importance estimation. However, the associated memory overhead is also negligible.

We further analyze the computational overhead introduced by each component of L2M. On 30-second video generation, L2M increases the inference time from $78.55$\,s to $79.21$\,s, corresponding to an overhead of only $0.66$\,s ($0.84\%$), while peak GPU memory increases from $36.05$\,GB to $36.32$\,GB ($0.75\%$). More than $95\%$ of the additional runtime is attributable to router forwarding. The EMA score update and sink-set modification together require less than $0.1$\,s. This low overhead is achieved because the lightweight router is evaluated only at the first denoising timestep and its predictions are reused throughout the remaining sampling process.

Overall, our approach achieves improved performance with almost no additional cost in inference time or memory usage.

\begin{table}[h]
\centering
\caption{Inference cost comparison on 30-second video generation.}
\begin{tabular}{lcc}
\toprule
\textbf{Method} & \textbf{Memory (GB)} & \textbf{Inference Time (s)} \\
\midrule
Baseline & 36.05 & 78.55 \\
Ours & 36.32 & 79.21 \\
\bottomrule
\end{tabular}
\end{table}

\paragraph{Effect of EMA coefficient $\alpha$}
We investigate the effect of the EMA coefficient $\alpha$ in Eq.~\ref{eq:ema}. A smaller $\alpha$ gives more weight to historical importance scores, resulting in more stable but less adaptive long-term memory selection. In contrast, a larger $\alpha$ emphasizes recent observations, making the sink set more responsive to changing content but also more prone to instability.

\begin{figure}[h]
  \centering
  \begin{minipage}{0.48\linewidth}
    \centering
    \includegraphics[width=\linewidth]{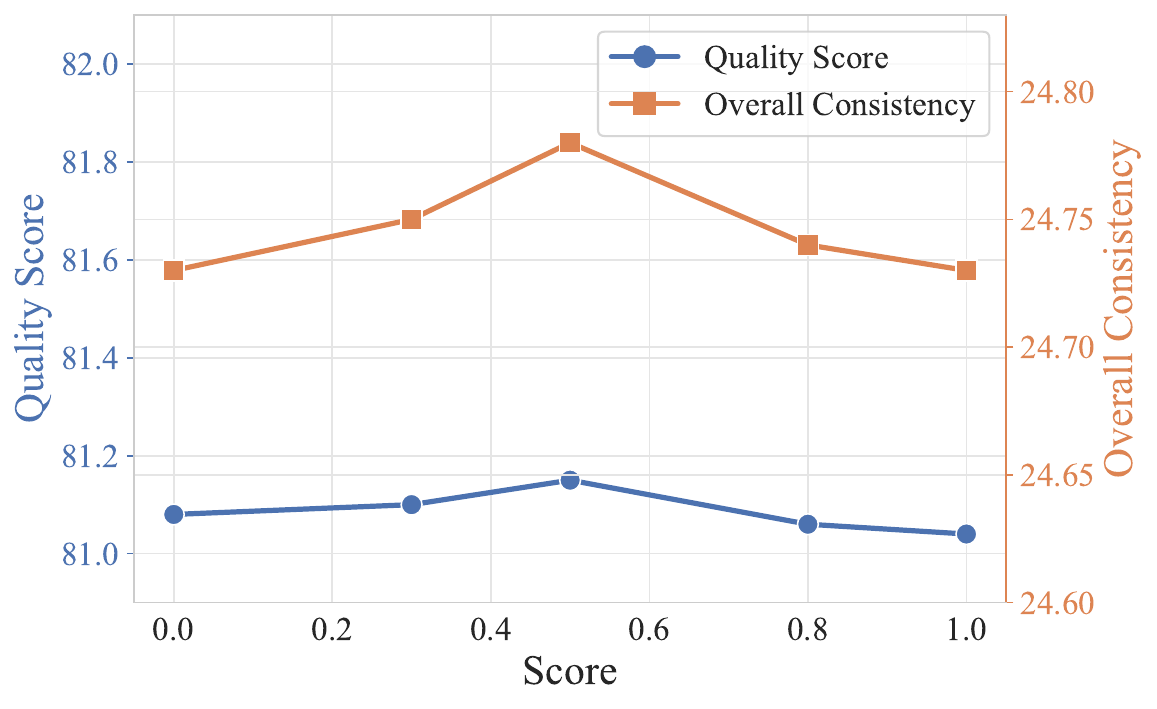}
    \caption{\textbf{Comparison of EMA coefficient $\alpha$.}
    Results on 30-second videos under different $\alpha$.}
    \label{fig:alpha_sweep}
  \end{minipage}
  \hfill
  \begin{minipage}{0.48\linewidth}
    \centering
    \includegraphics[width=\linewidth]{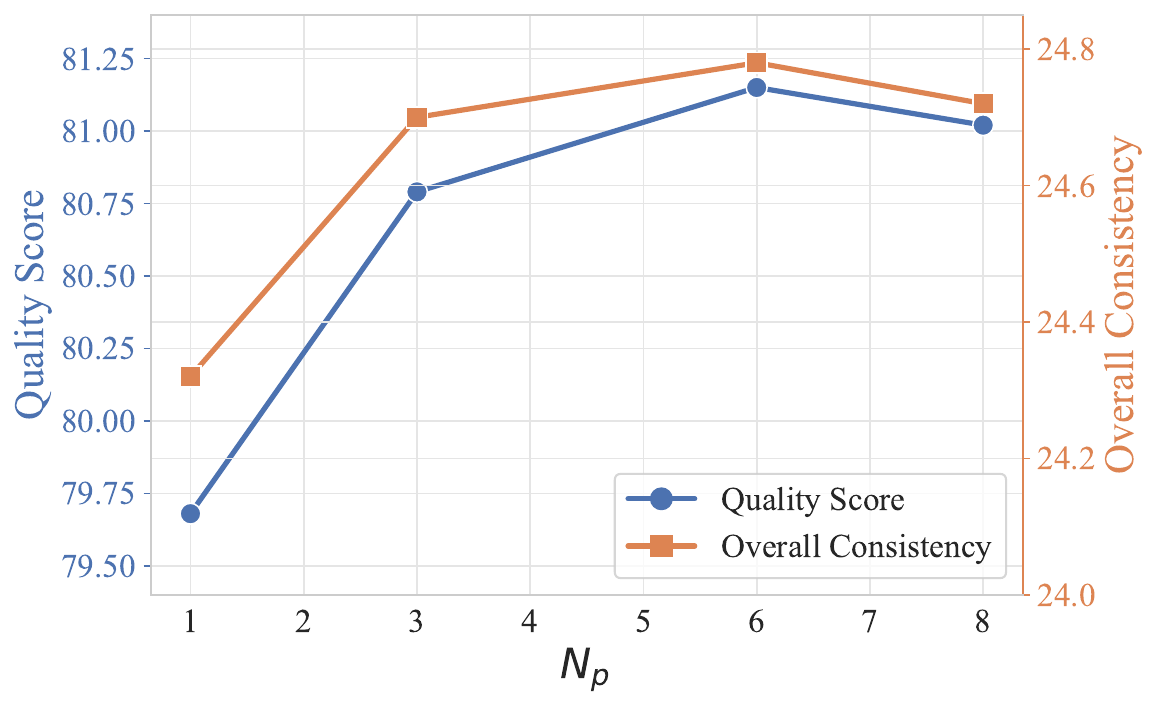}
    \caption{\textbf{Comparison of update interval $N_p$.}
    Results on 30-second videos under different $N_p$.}
    \label{fig:np_sweep}
  \end{minipage}
\end{figure}

As shown in Figure~\ref{fig:alpha_sweep}, a moderate value of $\alpha$ offers the best trade-off between stability and adaptability. In particular, $\alpha=0.5$ achieves the best performance, with the highest Quality Score and Overall Consistency. Both smaller and larger values lead to slight performance drops, suggesting that balanced temporal smoothing is important for effective dynamic memory updating.

\paragraph{Comparison of Long-term Memory Update Interval $N_p$}
We further study the long-term memory update interval $N_p$, which controls the lifespan of frames when updating the long-term memory. An overly small $N_p$ leads to excessively frequent memory updates, causing the model to rely on importance scores computed over short temporal windows that may not accurately reflect a frame's long-term contribution to video quality. In contrast, an overly large $N_p$ delays memory updates, resulting in stale context that cannot adapt to scene evolution in a timely manner. As shown in Figure~\ref{fig:np_sweep}, $N_p = 6$ achieves the best performance, striking a balance between stability and adaptability in long-term memory updating.

\begin{figure}[h]
  \centering
  \begin{minipage}{0.48\linewidth}
    \centering
    \includegraphics[width=\linewidth]{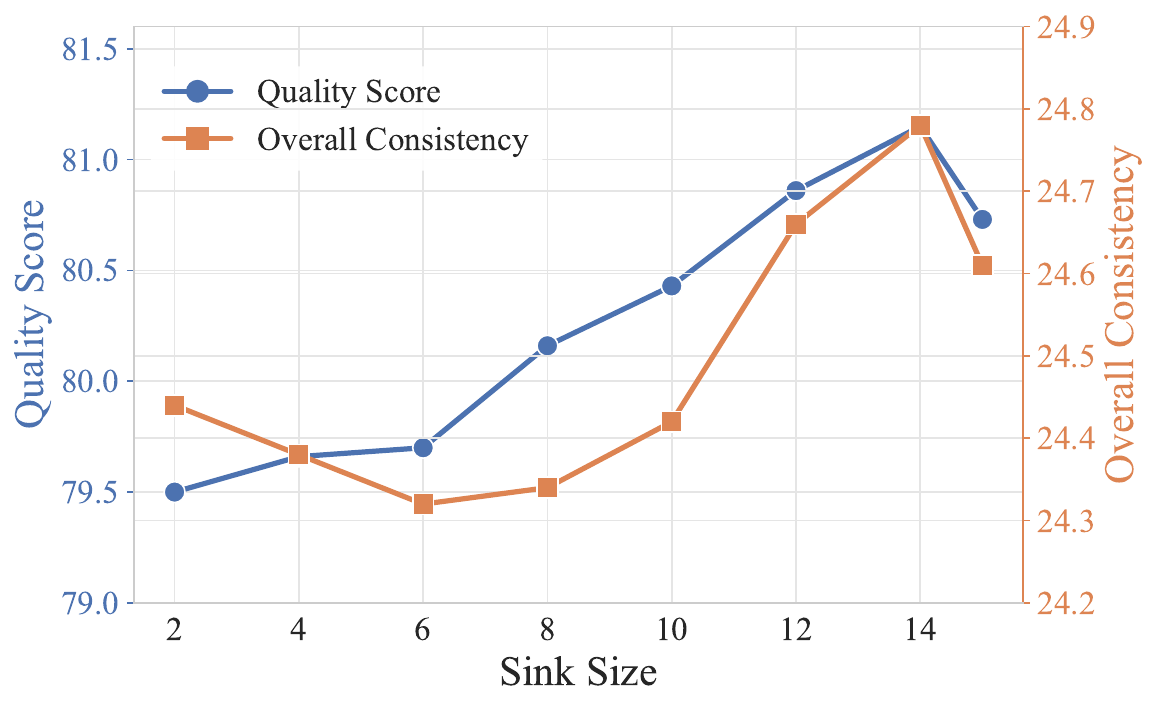}
    \caption{\textbf{Comparison of sink size.}
    Results on 30-second videos under different sink size.}
    \label{fig:sink_sweep}
  \end{minipage}
  \hfill
  \begin{minipage}{0.48\linewidth}
    \centering
    \includegraphics[width=\linewidth]{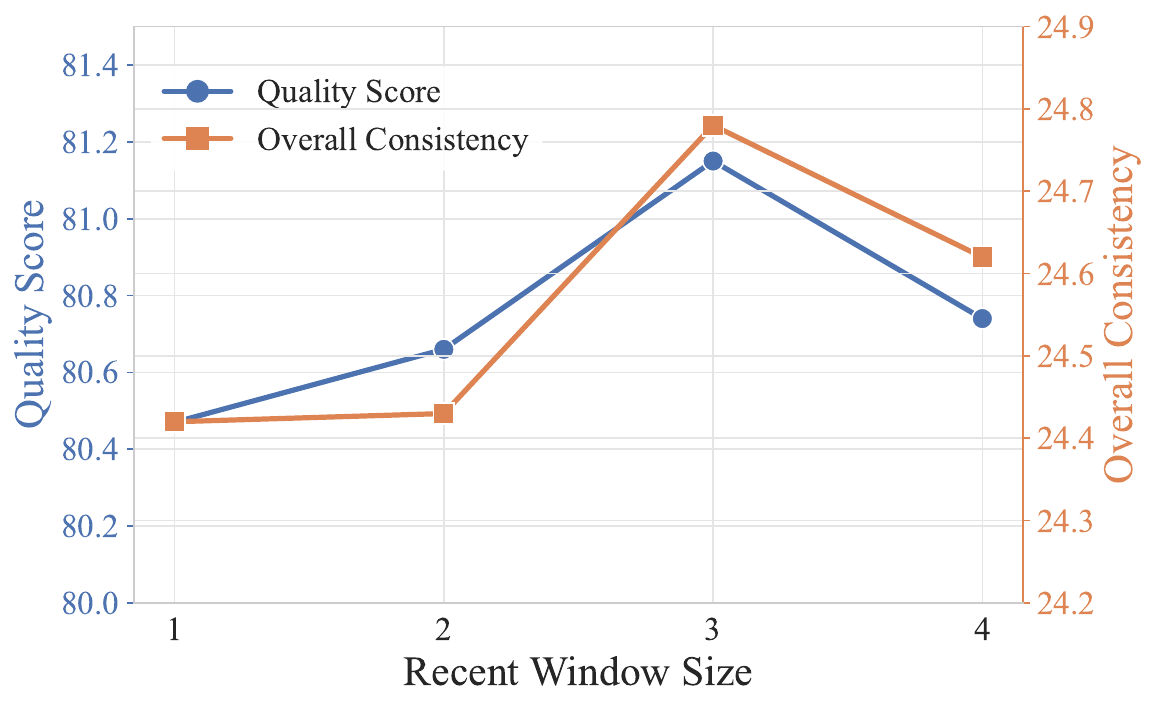}
    \caption{\textbf{Comparison of recent window size.}
    Results on 30-second videos under different recent window size.}
    \label{fig:recent_sweep}
  \end{minipage}
\end{figure}

\paragraph{Comparison of Sink Set Size}
We evaluate the effect of sink set size on model performance, as shown in Figure~\ref{fig:sink_sweep}. We observe that performance nearly saturates at a sink size of 14 (our default setting). Further decreasing the sink size leads to a gradual performance drop, while increasing it does not bring any improvement and may slightly degrade performance. This is mainly because a larger sink size forces more redundant frames to be preserved as long-term memory, reducing the efficiency of context utilization.

\paragraph{Comparison of Recent Window Size}
In our method, we preserve the three most recent frames as contextual frames during each eviction step, and only perform pruning on frames excluding the recent and sink frames. We further study the effect of different recent window lengths $|R|$. As shown in Figure~\ref{fig:recent_sweep}, we find that $|R|=3$ yields the best performance. Reducing $|R|$ leads to a noticeable performance drop, likely due to the disruption of temporal continuity in the video. This observation is also consistent with the results in Figure~\ref{fig:norm_router_prediction_pattern}, where the most recent three frames consistently receive the highest importance scores.

\paragraph{Comparison of Cache Budget}
We further analyze the sensitivity of L2M to the cache budget, using $K \in \{18,21,24,27,30\}$. As shown in Table~\ref{tab:cache_budget}, L2M achieves the highest Quality Score of $81.15$ at $K=21$. Overall Consistency improves modestly with larger cache budgets, increasing from $24.72$ at $K=18$ to $24.84$ at $K=30$. However, the score at $K=21$ is already within $0.06$ of this maximum while using $30\%$ fewer cache slots. Because increasing $K$ also raises KV-cache memory consumption and attention computation, we adopt $K=21$ as the default setting, which provides the best trade-off between efficiency and generation quality. These results also demonstrate that the effectiveness of L2M does not rely on an excessively large context budget.

\begin{table}[t]
  \centering
  \caption{\textbf{Comparison of cache budget $K$.}
  Higher is better for both metrics. The best results are in \textbf{bold}, and the second-best results are \underline{underlined}.}
  \label{tab:cache_budget}
  \small
  \begin{tabular}{ccc}
    \toprule
    \textbf{Cache Budget ($K$)} &
    \textbf{Quality Score} $\uparrow$ &
    \textbf{Overall Consistency} $\uparrow$ \\
    \midrule
    18 & 80.98             & 24.72 \\
    \rowcolor{black!2}
    21 & \textbf{81.15}    & 24.78 \\
    24 & 81.05             & 24.79 \\
    27 & \underline{81.10} & \underline{24.83} \\
    30 & 81.02             & \textbf{24.84} \\
    \bottomrule
  \end{tabular}
\end{table}

\paragraph{Comparison of Degradation Strength.}
We further investigate the sensitivity of L2M to the degradation strength used during training. As shown in Table~\ref{tab:degradation_sensitivity}, removing degradation-aware training reduces the Quality Score and Overall Consistency to $80.66$ and $24.68$, respectively. Applying moderate perturbations with the default setting ($p=0.2$, $\beta=0.1$) yields the best results, reaching $81.15$ and $24.78$. Increasing the perturbation strength beyond this setting slightly degrades performance to $80.92$ and $24.73$. These results suggest that moderate degradation provides effective robustness regularization, whereas overly strong perturbations may distort historical features and hinder importance estimation.

\begin{table}[t]
  \centering
  \caption{\textbf{Comparison of degradation strength.}
  Higher is better for both metrics. The best results are in \textbf{bold}.}
  \label{tab:degradation_sensitivity}
  \small
  \begin{tabular}{lcc}
    \toprule
    Training Setting &
    \textbf{Quality Score} $\uparrow$ &
    \textbf{Overall Consistency} $\uparrow$ \\
    \midrule
    No degradation ($p=0,\beta=0$)        & 80.66          & 24.68          \\
    \rowcolor{black!2}
    Default ($p=0.2,\beta=0.1$)           & \textbf{81.15} & \textbf{24.78} \\
    Stronger degradation                  & 80.92          & 24.73          \\
    \bottomrule
  \end{tabular}
\end{table}

\paragraph{Frame-wise versus Token-wise Eviction}
In this paper, we find that using frames as the minimal unit for context retention and eviction is more reasonable than using tokens. Specifically, we compare against a token-wise eviction strategy and obtain a quality score of 80.45 on 30-second videos, which is significantly lower than our frame-wise method (81.15). This degradation mainly arises because token-wise eviction introduces large temporal gaps in positional encodings across different frames. Although applying temporal RoPE adjustment can partially compensate for this issue, it forces tokens from different frames to be mapped into overly similar temporal positions, which in turn confuses the model’s contextual understanding and leads to inferior generation quality.

\begin{figure}[t]
    \centering
    \begin{subfigure}[t]{0.24\textwidth}
        \centering
        \includegraphics[width=\linewidth]{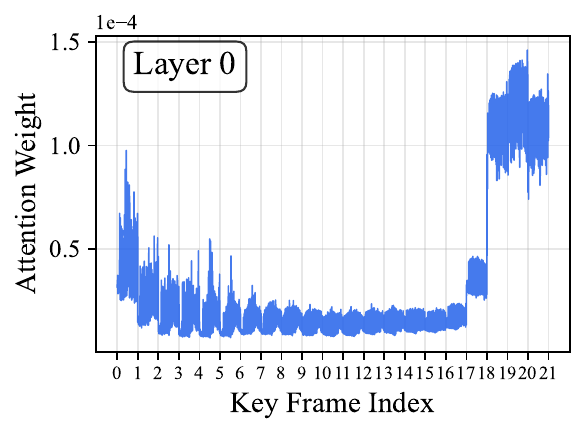}
    \end{subfigure}
    \hfill
    \begin{subfigure}[t]{0.24\textwidth}
        \centering
        \includegraphics[width=\linewidth]{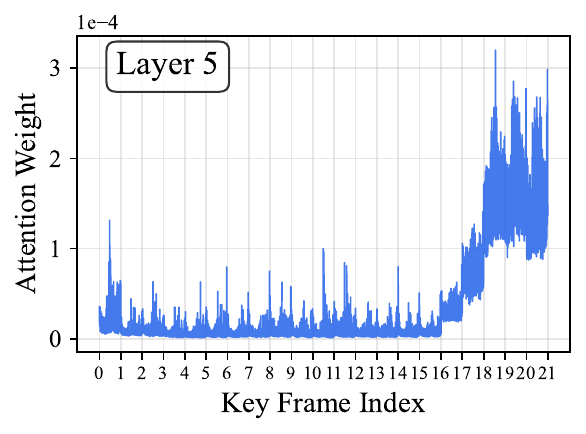}
    \end{subfigure}
    \hfill
    \begin{subfigure}[t]{0.24\textwidth}
        \centering
        \includegraphics[width=\linewidth]{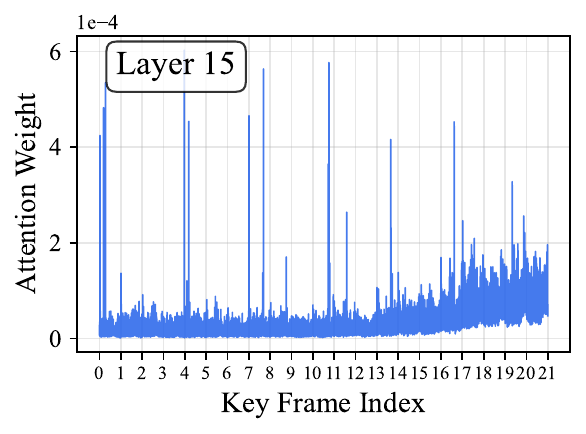}
    \end{subfigure}
    \hfill
    \begin{subfigure}[t]{0.24\textwidth}
        \centering
        \includegraphics[width=\linewidth]{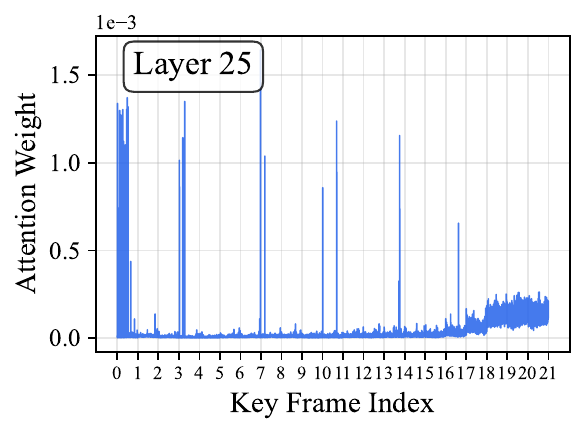}
    \end{subfigure}

    \caption{Visualization of attention weights across layers. Query-averaged attention maps illustrate how each key frame is attended during the generation of the final chunk.}
    \label{fig:attention_weights}
\end{figure}

\paragraph{Attention Pattern Analysis}
In this work, we observe that attention weight distributions vary significantly across different samples and network layers. Unlike prior observations in StreamLLM \cite{xiao2023efficient}, where attention responses are predominantly concentrated at the beginning of the sequence, we find that high-attention regions can emerge at different positions depending on the sample and the layer, as illustrated in Figure~\ref{fig:attention_weights}. 

This suggests that, in autoregressive video generation, sink tokens are not necessarily confined to the initial positions of the sequence, but may instead be distributed across the entire context. In practice, earlier frames and nearby frames often exhibit higher relevance, which explains their generally stronger attention responses. However, we also observe clear discrepancies in high-response regions across layers (e.g., between layer 15 and layer 25), indicating that attention patterns are both content-dependent and layer-specific.

These findings motivate the use of a content-aware, dynamically adaptive sink mechanism, rather than a position-based static sink.

\paragraph{Consistency of Importance Prediction across Diffusion Timesteps}
In our design, the importance Router is evaluated only at the first timestep, rather than being recomputed at every timestep. The predicted frame-wise importance pattern is then reused throughout subsequent steps. This design enables immediate token eviction when a new frame begins to be generated, thereby avoiding additional memory overhead. It also eliminates the extra computational cost associated with repeatedly running the Router at each timestep.

\begin{figure}
    \centering
    \includegraphics[width=1.0\linewidth]{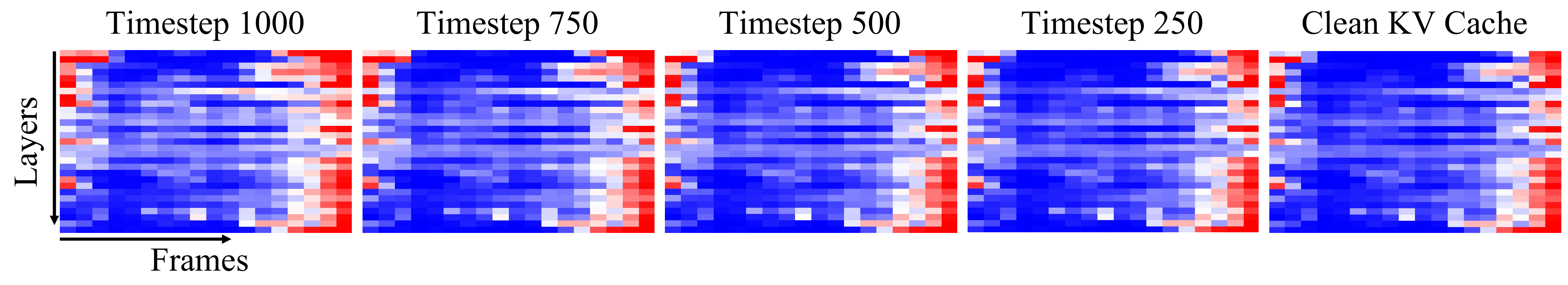}
    \caption{Consistency of importance scores across diffusion timesteps. The stable importance patterns observed at different timesteps (1000, 750, 500, 250, and the final clean KV cache) indicate that predictions made by the Router at the initial timestep ($t=1000$) remain reliable throughout the entire denoising process.}
    \label{fig:router_consistency}
\end{figure}

As shown in Figure \ref{fig:router_consistency}, we verify that the importance score patterns predicted from features at different timesteps are highly consistent, and lead to comparable final video quality. This robustness mainly stems from our training strategy, where the Router is exposed to noisy features from randomly sampled timesteps. As a result, the Router learns to make reliable decisions under varying noise levels. Consequently, the importance scores predicted at the first timestep generalize well across the entire sampling process.

\paragraph{Additional Context Pruning}
Moreover, as shown in Figure~\ref{fig:router_consistency}, we observe that the magnitude of token importance scores gradually decreases as the diffusion process progresses. This trend suggests that tokens at later timesteps tend to be more redundant. 
This observation is consistent with prior findings that early timesteps primarily determine the coarse structure and semantic content of the generated frames, while later timesteps focus on refining fine-grained details. Consequently, redundancy is typically higher in the later stages of the diffusion process.
Motivated by this insight, we further explore reducing the attention window size at subsequent timesteps. Our results show that removing an additional 10\% of context tokens in later steps results in only a marginal performance drop of 0.3\%, while further reducing computational cost.

\begin{table}[h]
\centering
\caption{Results of additional context pruning at later timesteps.}
\begin{tabular}{lcc}
\toprule
\textbf{Context Pruning} & \textbf{Quality Score} & \textbf{Overall Consistency}\\
\midrule
L2M & 81.15  & 24.78 \\
Uniform pruning (10\%) & 80.90 & 24.72 \\
Uniform pruning (20\%) & 80.68 & 24.67 \\
\bottomrule
\end{tabular}
\end{table}

\begin{figure}[h]
    \centering
    \includegraphics[width=1.0\linewidth]{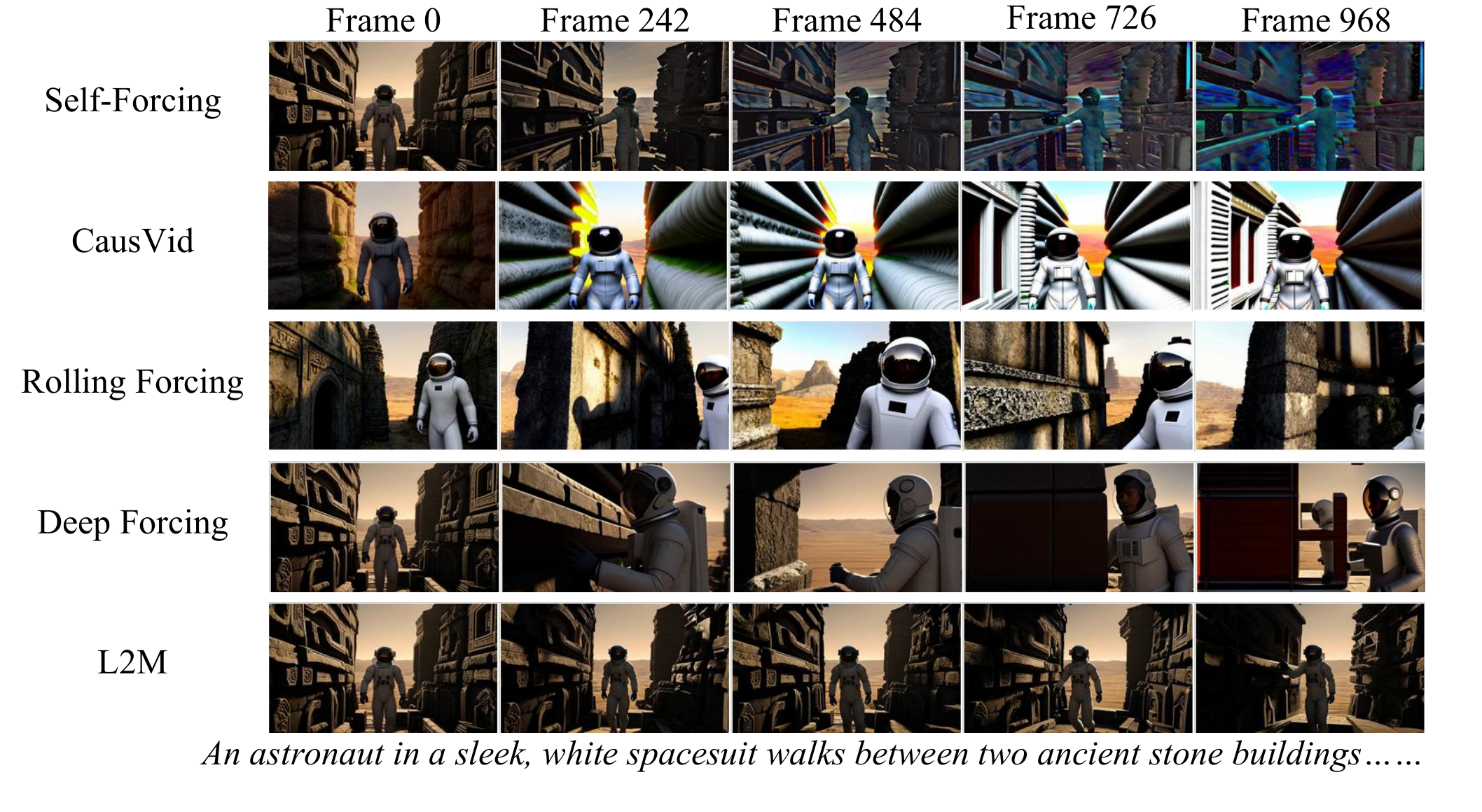}
    \caption{Qualitative comparison with existing methods. Our method achieves better visual quality, overall consistency patterns.}
    \label{fig:vis_2}
\end{figure}

\paragraph{Additional Comparisons with Existing Methods}
We provide additional visual comparisons with existing methods, as shown in Figure~\ref{fig:vis_2}. Our results exhibit better overall consistency and richer motion dynamics.

\paragraph{Router Reliability and Error Analysis}
Incorrect routing decisions may propagate errors to subsequent frames. To reduce its dependence on individual predictions, L2M incorporates three complementary safeguards: (1) degradation-aware training teaches the router to identify and downweight unreliable historical features; (2) EMA smoothing aggregates importance scores over time, reducing sensitivity to transient prediction noise; and (3) the $N_p$-step survival criterion prevents newly generated frames from being prematurely promoted to long-term memory.

We directly assess router reliability through a controlled diagnostic in which the current query feature and context positions are fixed. The router assigns average retention scores of $0.61$, $0.24$, and $0.19$ to clean, synthetically degraded, and rollout-degraded frames, respectively. The substantially lower scores for both types of degraded frames demonstrate that the router learns to suppress not only synthetic perturbations encountered during training but also artifacts arising from actual autoregressive rollouts.

Nevertheless, occasional routing errors may still propagate, particularly when the historical context contains semantic inconsistencies such as identity drift. Although the proposed mechanisms reduce this risk, they cannot eliminate it entirely.

\paragraph{Failure Cases}
Our method does not require additional training of the base model, enabling efficient and low-cost performance improvements. However, this design also limits the upper bound of achievable gains. Although L2M consistently outperforms existing baselines, it may still suffer from abrupt content changes in scenes with drastic motion or rapid scene transitions. In addition, since our approach operates as a dynamic memory regulation mechanism, poor-quality frames stored in the initial memory cache may not be evicted promptly and can propagate erroneous information to subsequent generation, thereby degrading future visual quality.

\begin{figure}[h]
    \centering
    \includegraphics[width=0.9\linewidth]{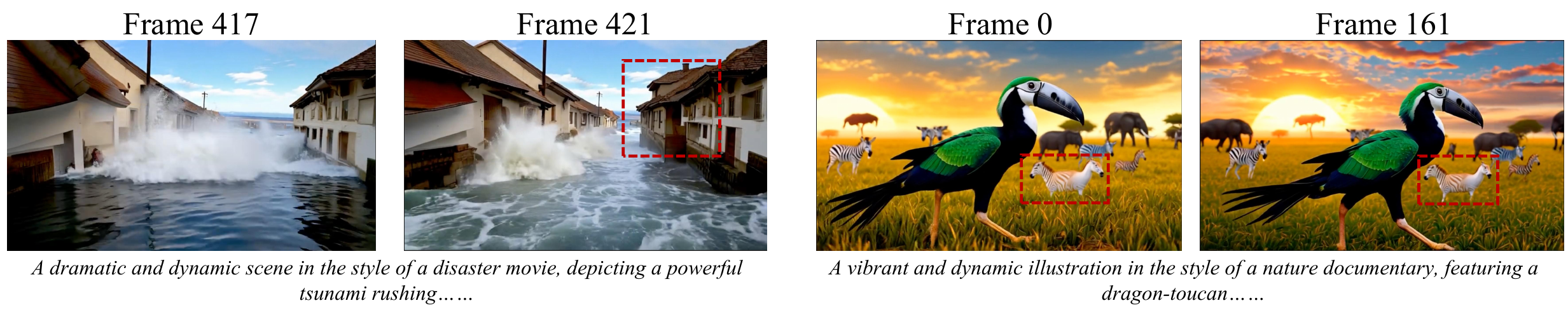}
    \caption{\textbf{Failure cases of L2M.} In the first example, rapid camera movement and highly dynamic scene changes cause abrupt content inconsistency, where a new house artifact appears after the waves sweep across the scene. In the second example, an incorrectly generated two-headed zebra appears in several initial frames and remains in subsequent generations due to delayed memory eviction.}
    \label{fig:failure_case}
\end{figure}

\clearpage
\section*{NeurIPS Paper Checklist}

\begin{enumerate}

\item {\bf Claims}
    \item[] Question: Do the main claims made in the abstract and introduction accurately reflect the paper's contributions and scope?
    \item[] Answer: \answerYes{} 
    \item[] Justification: The abstract and introduction clearly describe our main contributions, including dynamic long-term memory management, importance prediction router, and extensive evaluations on long video generation benchmarks. The claims are consistent with experimental results presented in the main paper.


\item {\bf Limitations}
    \item[] Question: Does the paper discuss the limitations of the work performed by the authors?
    \item[] Answer: \answerYes{} 
    \item[] Justification: We discuss limitations in the Appendix \ref{sec:limitation}.


\item {\bf Theory assumptions and proofs}
    \item[] Question: For each theoretical result, does the paper provide the full set of assumptions and a complete (and correct) proof?
    \item[] Answer: \answerNA{} 
    \item[] Justification: The paper does not include theoretical proofs. It focuses on empirical methods and experimental validation.


    \item {\bf Experimental result reproducibility}
    \item[] Question: Does the paper fully disclose all the information needed to reproduce the main experimental results of the paper to the extent that it affects the main claims and/or conclusions of the paper (regardless of whether the code and data are provided or not)?
    \item[] Answer: \answerYes{} 
    \item[] Justification: We provide detailed model architecture, training settings, inference procedure, and evaluation protocol, sufficient to reproduce the main results.


\item {\bf Open access to data and code}
    \item[] Question: Does the paper provide open access to the data and code, with sufficient instructions to faithfully reproduce the main experimental results, as described in supplemental material?
    \item[] Answer: \answerYes{} 
    \item[] Justification:  
     The code is provided in the attach file.


\item {\bf Experimental setting/details}
    \item[] Question: Does the paper specify all the training and test details (e.g., data splits, hyperparameters, how they were chosen, type of optimizer) necessary to understand the results?
    \item[] Answer: \answerYes{} 
    \item[] Justification: The paper includes full training details, hyperparameters, data processing, and evaluation settings in the main paper and appendix.


\item {\bf Experiment statistical significance}
    \item[] Question: Does the paper report error bars suitably and correctly defined or other appropriate information about the statistical significance of the experiments?
    \item[] Answer: \answerNA{} 
    \item[] Justification: We follow standard evaluation protocols for video generation benchmarks and report averaged results, variance across runs is small and not the primary focus of this work.


\item {\bf Experiments compute resources}
    \item[] Question: For each experiment, does the paper provide sufficient information on the computer resources (type of compute workers, memory, time of execution) needed to reproduce the experiments?
    \item[] Answer: \answerYes{} 
    \item[] Justification: We provide details on training hardware (GPU type, number of devices), training time, and inference cost in the appendix.

    
\item {\bf Code of ethics}
    \item[] Question: Does the research conducted in the paper conform, in every respect, with the NeurIPS Code of Ethics \url{https://neurips.cc/public/EthicsGuidelines}?
    \item[] Answer: \answerYes{} 
    \item[] Justification: The work complies with the NeurIPS Code of Ethics, involving no human subjects or sensitive data.


\item {\bf Broader impacts}
    \item[] Question: Does the paper discuss both potential positive societal impacts and negative societal impacts of the work performed?
    \item[] Answer: \answerYes{} 
    \item[] Justification:  We discuss potential positive impacts in improving long video generation and possible negative impacts such as misuse for deepfake generation, along with mitigation considerations.

    
\item {\bf Safeguards}
    \item[] Question: Does the paper describe safeguards that have been put in place for responsible release of data or models that have a high risk for misuse (e.g., pre-trained language models, image generators, or scraped datasets)?
    \item[] Answer: \answerNA{} 
    \item[] Justification: The paper does not release high-risk generative models with direct deployment safeguards beyond standard research usage.


\item {\bf Licenses for existing assets}
    \item[] Question: Are the creators or original owners of assets (e.g., code, data, models), used in the paper, properly credited and are the license and terms of use explicitly mentioned and properly respected?
    \item[] Answer: \answerYes{} 
    \item[] Justification: We use publicly available datasets and pretrained models with properly cited sources and respect their licenses as stated in the paper.


\item {\bf New assets}
    \item[] Question: Are new assets introduced in the paper well documented and is the documentation provided alongside the assets?
    \item[] Answer: \answerNA{} 
    \item[] Justification: The paper does not introduce new datasets or external assets beyond the proposed method and model.


\item {\bf Crowdsourcing and research with human subjects}
    \item[] Question: For crowdsourcing experiments and research with human subjects, does the paper include the full text of instructions given to participants and screenshots, if applicable, as well as details about compensation (if any)? 
    \item[] Answer:  \answerNA{} 
    \item[] Justification: The paper does not involve human subject experiments or crowdsourcing studies.


\item {\bf Institutional review board (IRB) approvals or equivalent for research with human subjects}
    \item[] Question: Does the paper describe potential risks incurred by study participants, whether such risks were disclosed to the subjects, and whether Institutional Review Board (IRB) approvals (or an equivalent approval/review based on the requirements of your country or institution) were obtained?
    \item[] Answer: \answerNA{} 
    \item[] Justification: The paper does not involve human subjects.

\item {\bf Declaration of LLM usage}
    \item[] Question: Does the paper describe the usage of LLMs if it is an important, original, or non-standard component of the core methods in this research? Note that if the LLM is used only for writing, editing, or formatting purposes and does \emph{not} impact the core methodology, scientific rigor, or originality of the research, declaration is not required.
    \item[] Answer: \answerYes{} 
    \item[] Justification: A large language model (Qwen2.5-7B-Instruct) is used only for prompt refinement and does not influence the core method design or evaluation.

\end{enumerate}

\end{document}